\documentclass{article} 
\usepackage{iclr2027_conference,times}

\usepackage{amsmath,amsfonts,bm}

\def\eqref#1{equation~\ref{#1}}

\def\1{\bm{1}}

\DeclareMathAlphabet{\mathsfit}{\encodingdefault}{\sfdefault}{m}{sl}
\SetMathAlphabet{\mathsfit}{bold}{\encodingdefault}{\sfdefault}{bx}{n}

\usepackage{hyperref}
\usepackage{url}
\usepackage{booktabs}
\usepackage{verbatim}

\usepackage[ruled]{algorithm2e} 

\SetAlFnt{\small}
\SetAlCapFnt{\small}
\SetAlCapNameFnt{\small}
\SetAlCapHSkip{0pt}

\usepackage{float}

\usepackage[capitalize]{cleveref}
\usepackage{enumitem}
\usepackage{xcolor}
\usepackage[table]{xcolor}
\usepackage{pifont}
\usepackage{kotex}
\newcommand{\cmark}{\textcolor{green}{\ding{51}}}
\newcommand{\xmark}{\textcolor{red}{\ding{55}}}

\usepackage{tcolorbox}
\usepackage{cuted}
\usepackage{makecell}
\usepackage{multirow}
\usepackage{wrapfig}

\usepackage{fontawesome5}

\def\DatasetName{TriLayer} 
\def\MethodName{DBL-Diffusion}
\def\MethodNameInsert{DBL-Insert}
\def\MethodNameDecompose{DBL-Decompose}
\def\MethodNameRefinement{AGBI}

\definecolor{bananayellow}{rgb}{1.0, 0.88, 0.21}
\definecolor{forestgreen(web)}{rgb}{0.13, 0.55, 0.13}

\newcommand{\red}[1]{{\color{red}#1}}
\newcommand{\blue}[1]{{\color{blue}#1}}

\newcommand{\green}[1]{{\color{forestgreen(web)}#1}}

\title{Explicit Layer Modeling for Video Object Insertion and Video Layer Decomposition}

\iclrfinalcopy

\author{Kyujin Han, Seungjoo Shin, Sunghyun Cho$^{\dagger}$ \\
POSTECH\\
\texttt{\{kyujin,seungjoo.shin,s.cho\}@postech.ac.kr} \\
$^{\dagger}$ Corresponding author
}

\begin{document}

\maketitle

\vspace{-2ex}
\begin{center}
  \href{https://kyujinhan.github.io/dual-branch-layered-diffusion/}{\faGithub\ \blue{Project Page}}
\end{center}

\begin{figure}[h]
    \centering
    \includegraphics[width=\textwidth]{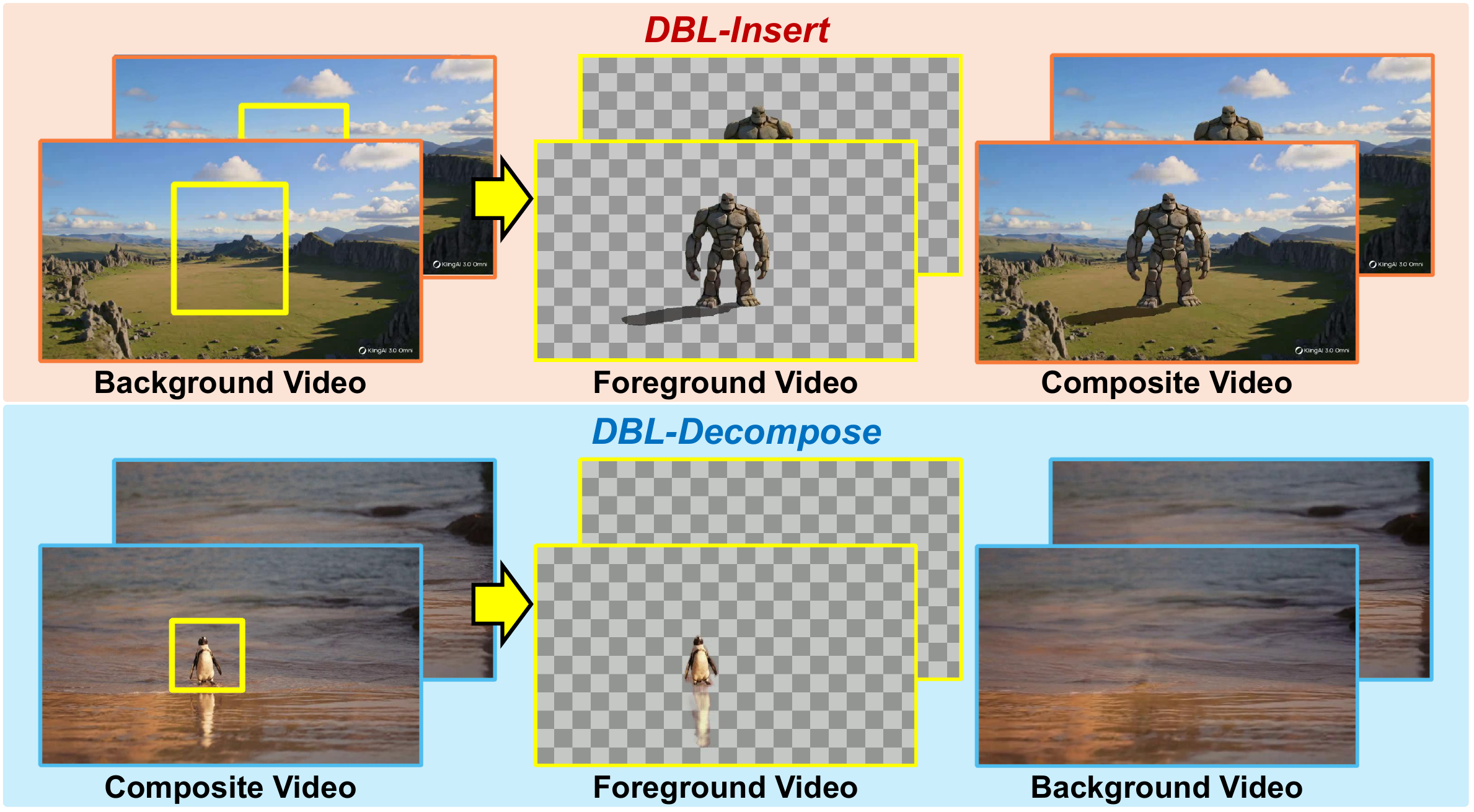}
    \vspace{-5ex}
    \caption{
    Overview of our proposed \MethodName{}. Our framework addresses layered video tasks through explicit layer modeling and consists of two instantiations: \MethodNameInsert{} for video layered object insertion and \MethodNameDecompose{} for video layer decomposition. The yellow box denotes the input mask.
    }
    \label{fig:teaser}
\end{figure}

\begin{abstract}
Most video editing systems still lack explicit layered video representations, limiting realistic compositing, object reuse, and consistent manipulation.
This limitation is particularly evident in video object insertion and video layer decomposition, where existing methods lack direct supervision for foreground layers that capture both objects and their associated visual effects.
We introduce \DatasetName{}, a triplet video dataset containing aligned composite--background--foreground videos, where the foreground layers include both object appearance and associated visual effects. 
With aligned triplet supervision, \DatasetName{} enables explicit supervised learning of layered video representations for the first time.
Building on this dataset, we propose \MethodName{}, a dual-branch diffusion framework that jointly models scene-level RGB content and RGBA foreground layers through cross-branch interaction during denoising.
We instantiate the framework in two tasks: \MethodNameInsert{} for layered object insertion, which generates explicit RGBA layers for realistic compositing and flexible post-editing, and \MethodNameDecompose{} for video layer decomposition, which recovers foreground and background layers using triplet supervision. 
Experiments demonstrate that explicit layer modeling substantially improves both insertion fidelity and decomposition quality.
\end{abstract}

\section{Introduction}


Video editing for film, advertising, and digital content creation often involves inserting new objects into existing footage or extracting objects for reuse in different compositions.
Representing objects and backgrounds as separate layers facilitates these operations by allowing individual elements to be modified and recomposed independently.
This flexibility motivates two complementary tasks: \emph{video object insertion}, which integrates new objects into a background video, and \emph{video layer decomposition}, which separates an existing video into foreground and background layers.
Generating and recovering editable foreground layers for these tasks can provide greater control over both the initial composition and subsequent edits.

Early approaches to \emph{video object insertion} insert objects into background videos through inpainting~\citep{mou2024revideo, bian2025videopainter, jiang2025vace, tu2025videoanydoor}, but often alter nearby background content and offered limited flexibility for post-editing. 
More recent approaches~\citep{zi2025se, chen2025omniinsert, jin2025insertanywhere} improve object insertion quality by learning from paired background--composite videos. 
However, these datasets do not provide explicit foreground layers that include both the objects and their associated visual effects, such as shadows and reflections.
Consequently, existing methods learn to generate the inserted object and its effects within the background-integrated RGB video, rather than as an independently composable foreground layer, which can lead to unnecessary background changes and limit independent post-editing of the inserted object and its associated effects.
Similarly, \emph{video layer decomposition} seeks to recover foreground and background layers from a composite video, but existing methods do not learn from explicit foreground-layer supervision.
Early \emph{omnimatte} approaches predict foreground layers through per-scene optimization~\citep{lu2021omnimatte, lin2023omnimatterf, suhail2023omnimatte3d}, resulting in limited generalization across in-the-wild scenes.
Recent methods~\citep{lee2025generative, samuel2025omnimattezero} improve reconstruction quality, but still infer RGBA foreground layers through implicit inference without explicit foreground-layer ground truth, leading to incomplete visual effect separation or overly translucent foregrounds.

Although video object insertion and video layer decomposition originate from different goals, both tasks face the same fundamental bottleneck: the lack of direct supervision for learning foreground layers that explicitly capture objects and their associated visual effects.
Directly learning such layered representations requires spatially and temporally aligned composite, clean background, and effect-inclusive foreground layers, which existing video datasets do not provide.
We address this limitation by introducing \emph{\DatasetName{}}, a triplet video dataset consisting of aligned foreground--background--composite videos, and \emph{\MethodName{}} (Dual-Branch Layered Diffusion), a unified dual-branch diffusion framework that jointly models RGB video content and RGBA foreground layers.

To construct \emph{\DatasetName{}}, we design a multi-stage pipeline from in-the-wild videos. 
From each composite video, we reconstruct the background video by removing the object and obtain a foreground layer that captures the object and its associated visual effects. We then refine the reconstructed backgrounds using \emph{Appearance-guided background inpainting (AGBI)}, which reduces residual artifacts while preserving the background appearance.
The overall pipeline combines automated processing, VLM-based filtering, and human verification to construct high-quality layered supervision.
By providing composite–background–foreground correspondences, \emph{\DatasetName{}} enables supervised learning of layered video representations for the first time, filling a critical gap in existing datasets.

Building upon \DatasetName{}, we develop \emph{\MethodName{}} to jointly model scene-level RGB content and explicit RGBA foreground layers. To this end, \emph{\MethodName{}} employs dual diffusion branches that separately process an RGB video and an RGBA foreground layer, 
while enabling information exchange throughout the denoising process via cross-attention interaction. 
This joint modeling enables the model to generate both a scene-level RGB video and an explicit RGBA foreground layer that captures object-induced effects arising from their spatiotemporal interactions with the surrounding scene.
We instantiate this framework in two complementary tasks: (1) \emph{\MethodNameInsert{}} for video layered object insertion, which generates explicit RGBA foreground layers for realistic compositing and flexible post-editing—the first demonstration of layered object insertion in the video domain; and (2) \emph{\MethodNameDecompose{}} for video layer decomposition, which recovers foreground and background layers from composite videos using triplet supervision. Together, these two instantiations show that explicit layer modeling is key to both high-fidelity insertion and more accurate decomposition.

\clearpage
Our contributions are summarized as follows:
\begin{itemize}[leftmargin=0pt, itemsep=0pt, topsep=0pt, partopsep=0pt]
    \item We introduce \emph{\DatasetName{}}, a triplet video dataset that provides composite–background–foreground correspondences, enabling supervised learning of layered video representations for the first time. We further present a synthesis pipeline for scalable construction of high-quality layered video datasets.
    
    \item We develop \emph{\MethodName{}}, a unified dual-branch diffusion framework that jointly models RGB and RGBA representations through cross-branch interaction during denoising. This framework explicitly captures layered video structure and serves as a general backbone for multiple layered video tasks.

    \item We instantiate \emph{\MethodName{}} in two tasks: \emph{\MethodNameInsert{}} for video layered object insertion and \emph{\MethodNameDecompose{}} for video layer decomposition, achieving high-fidelity insertion and substantial improvements in decomposition.
\end{itemize}

\section{\DatasetName{} Dataset}
\label{label:dataset_sec}


\subsection{Dataset Overview}

\DatasetName{} provides
aligned composite, background, and foreground videos for each sample.
The composite video contains the original scene with the object present.
The foreground video and its alpha matte capture
opaque object regions and semi-transparent effects such as shadows and reflections. The background video contains neither the object nor its associated effects, serving as a clean reference for decomposition and as the input for layered object insertion.
Although each sample contains three aligned videos, these are not
independent layers; the composite is constructed by alpha-compositing
the foreground layer onto the background.
The dataset contains 3,908 video triplets covering diverse objects, motions, and environments. 

\cref{tab:dataset} compares \DatasetName{} with existing public datasets.
Video matting datasets~\citep{zhang2021attention, lin2021real, sun2021deep} focus on foreground extraction and do not include alpha mattes for visual
effects. Senorita-2M~\citep{zi2025se} and VPData~\citep{bian2025videopainter}
lack clean background videos, making layered decomposition impossible to supervise. ROSE++~\citep{jin2025insertanywhere} contains static
objects in virtual scenes and therefore cannot model dynamic
object--scene interactions. In contrast, \DatasetName{} offers complete layered supervision, enabling models to learn both object appearance and
object-induced visual effects from fully aligned composite, background,
and foreground.

\subsection{Dataset Construction Pipeline}

\begin{figure*}
    \vspace{-3pt}
    \centering
    \includegraphics[width=\textwidth]{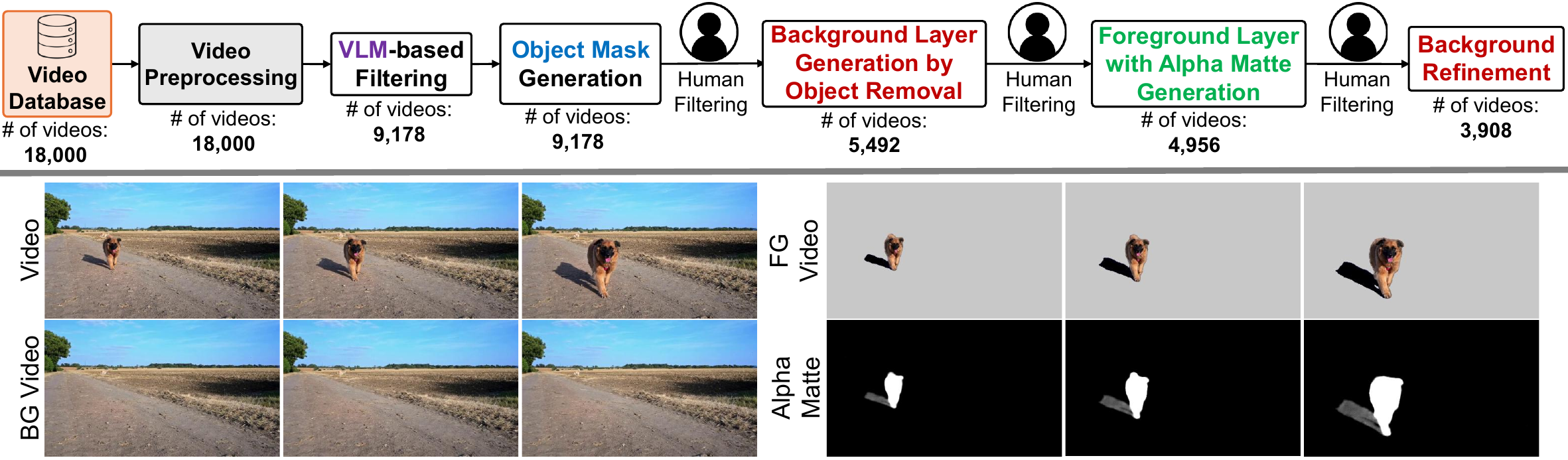}
    \vspace{-5ex}
    \caption{Detailed \DatasetName{} dataset construction pipeline and representative examples.
    }
    \label{fig:dataset}
\end{figure*}

\cref{fig:dataset} illustrates a multi-stage pipeline used to construct \DatasetName{}, which derives foreground--background--composite triplets from in-the-wild videos, ensuring high-quality layered supervision.

\paragraph{Video preprocessing and candidate selection.}
We begin by collecting diverse in-the-wild videos from the Pexels~\citep{pexels}. The raw videos are resized and temporally trimmed to match the diffusion training setup. For videos containing multiple shots, we detect shot boundaries using PySceneDetect
\citep{PysceneDetect} and retain only the first clean shot, as subsequent
shots often introduce transition artifacts (e.g., fade-outs or abrupt
cuts) that degrade foreground–background separation. We then apply the VLM~\citep{bai2025qwen25vltechnicalreport} to identify the main foreground object and filter
out clips with severe blur, heavy occlusion, or extremely small objects.

\paragraph{Background and foreground layer reconstruction.}
For each video sample, we first extract per-frame object masks using Grounded-SAM2~\citep{ren2024grounding, ravi2024sam}. These masks guide background reconstruction, where foreground regions are removed and inpainted using an off-the-shelf object removal model~\citep{miao2025rose}, yielding an initial background video.
Given the composite and reconstructed background, 
we then extract an RGBA
foreground layer that captures both object appearance and associated
visual effects. We combine complementary cues from Gen-Omnimatte~\citep{lee2025generative},
video matting (MatAnyone~\citep{yang2025matanyone}), and segmentation (Grounded-SAM2): Gen-Omnimatte provides an initial RGBA layer with reliable effect extraction, while MatAnyone and Grounded-SAM2 refine object
opacity and boundaries. The final alpha matte is the per-pixel maximum
across sources, and RGB values are selected from the source with the
winning alpha. This hybrid strategy improves both efficiency and quality.
Additional details (\cref{eq:dataset_eq}) are provided in the supplementary material.

\begin{table*}[t]
    \centering
    \caption{Comparison with existing open-source video matting and video object insertion datasets. 
    }    
    \resizebox{\textwidth}{!}{
        \begin{tabular}{lccccccccc}
        \toprule
        Dataset & 
        \makecell{Composition \\ Video} & 
        \makecell{BG Video \\ (Visual effect removed)} & 
        \makecell{0-1 \\ Mask} & 
        \makecell{Alpha \\ matte} & 
        \makecell{Alpha matte \\ (Visual effect included)} & 
        \makecell{Dynamic \\ Object} & 
        \makecell{Real-World \\ Video} & 
        \makecell{\# of \\ Frames} \\
        \midrule

        DVM~\citep{sun2021deep}
        & \cmark & \xmark & \xmark & \cmark & \xmark & \cmark & \cmark & 81$<$ \\

        VideoMatting108~\citep{zhang2021attention}
        & \cmark & \xmark & \xmark & \cmark & \xmark & \cmark & \cmark & 81$<$ \\

        VideoMatte240K~\citep{lin2021real}
        & \cmark & \xmark & \xmark & \cmark & \xmark & \cmark & \cmark & 81$<$ \\
        
        Senorita-2M~\citep{zi2025se}
        & \cmark & \xmark & \cmark & \xmark & \xmark & \cmark & \cmark & 33$\sim$64 \\
        
        VPData~\citep{bian2025videopainter}
        & \cmark & \xmark & \cmark & \xmark & \xmark & \cmark & \cmark & 81$<$ \\
        
        ROSE++~\citep{jin2025insertanywhere}
        & \cmark & \cmark & \cmark & \xmark & \xmark & \xmark & \xmark & 81$<$ \\
        
        \cellcolor{blue!15}\DatasetName{} (Ours) 
        & \cellcolor{blue!15}\cmark & \cellcolor{blue!15}\cmark & \cellcolor{blue!15}\cmark & \cellcolor{blue!15}\cmark & \cellcolor{blue!15}\cmark & \cellcolor{blue!15}\cmark & \cellcolor{blue!15}\cmark & \cellcolor{blue!15}81 \\
        
        \bottomrule
    \end{tabular}
    }
    \label{tab:dataset}
\end{table*}

\paragraph{Background refinement.}
Existing video object removal methods often leave residual artifacts, such as afterimages and ghosting, in the reconstructed background.
A straightforward solution is to refine these regions with a pretrained inpainting model using a binary mask derived from the foreground alpha matte.
However, as shown in \cref{fig:AGBI_introduce}, object-level masking can induce the model to generate unintended objects following the mask's shape, whereas bounding-box masking can change the surrounding background appearance.

To address these limitations, we propose \emph{\MethodNameRefinement{}}, which uses a pretrained video inpainting model to refine the initial background produced by the object removal model.
Our key idea is to use the object-removal result, which may contain residual artifacts, as an appearance guide for inpainting. We initialize the inpainting process from a partially noised version of this result, allowing the model to correct these artifacts while preserving the desired background appearance.
We further use a two-stage masking strategy to control this refinement.
During the initial denoising steps, a bounding-box mask avoids exposing the object's detailed silhouette as a shape cue that could encourage its regeneration.
We then switch to a precise object-level mask to restrict subsequent refinement to the target region and limit unnecessary changes to the surrounding background.
Given the source background video $X^{\mathrm{src}}$ obtained from the object removal model, we encode it into an initial latent $Z_0^{\mathrm{src}}$ and initialize the target latent at a noise level controlled by $\sigma_s \in [0,1]$. Specifically, for $T$ denoising steps, we define the starting step $T_s = T-\lfloor T\sigma_s \rfloor+1$ and initialize the target latent as: 
\begin{equation}
Z_{t_{T_s}}^{\mathrm{tar}}
=
(1-t_{T_s})Z_0^{\mathrm{src}}
+
t_{T_s}\epsilon,
\qquad
\epsilon\sim\mathcal{N}(0,I)
\end{equation}
Here, $t_i$ denotes the noise level at step $i \in [T_s,T]$. During denoising, we use a step-dependent mask, starting with a bounding-box mask $B$ and switching to a precise object-level mask $M$ after $N$ steps:
\begin{equation}
V_i =
\begin{cases}
v_{\theta}(Z_{t_i}^{\mathrm{tar}},t_i,B,P), & (i - T_s) < N, \\
v_{\theta}(Z_{t_i}^{\mathrm{tar}},t_i,M,P), & (i - T_s) \geq N,
\end{cases}
\end{equation}
where $P$ and $v_{\theta}$ denote the prompt and the pretrained inpainting model, respectively.
We construct the final background videos by refining the object-removal outputs with \MethodNameRefinement{}.
More details and qualitative results (Sec.~\ref{sec:AGBI} and \cref{fig:agbi_more_qual}) are provided in the supplementary material.

\paragraph{Filtering.}
Starting from roughly 18,000 candidates, we apply automated filtering and multiple rounds of human verification to ensure mask accuracy, background stability, and foreground fidelity, resulting in 3,908 high-quality triplets suitable for training layered video representations. We then generate object-centric captions using a VLM that describe both the object and its visual effects. 
More detailed descriptions and examples (Sec.~\ref{sec:dataset}) are provided in the supplementary material.

\begin{figure*}[t]
    \vspace{-3pt}
    \centering
    \includegraphics[width=0.9\textwidth]{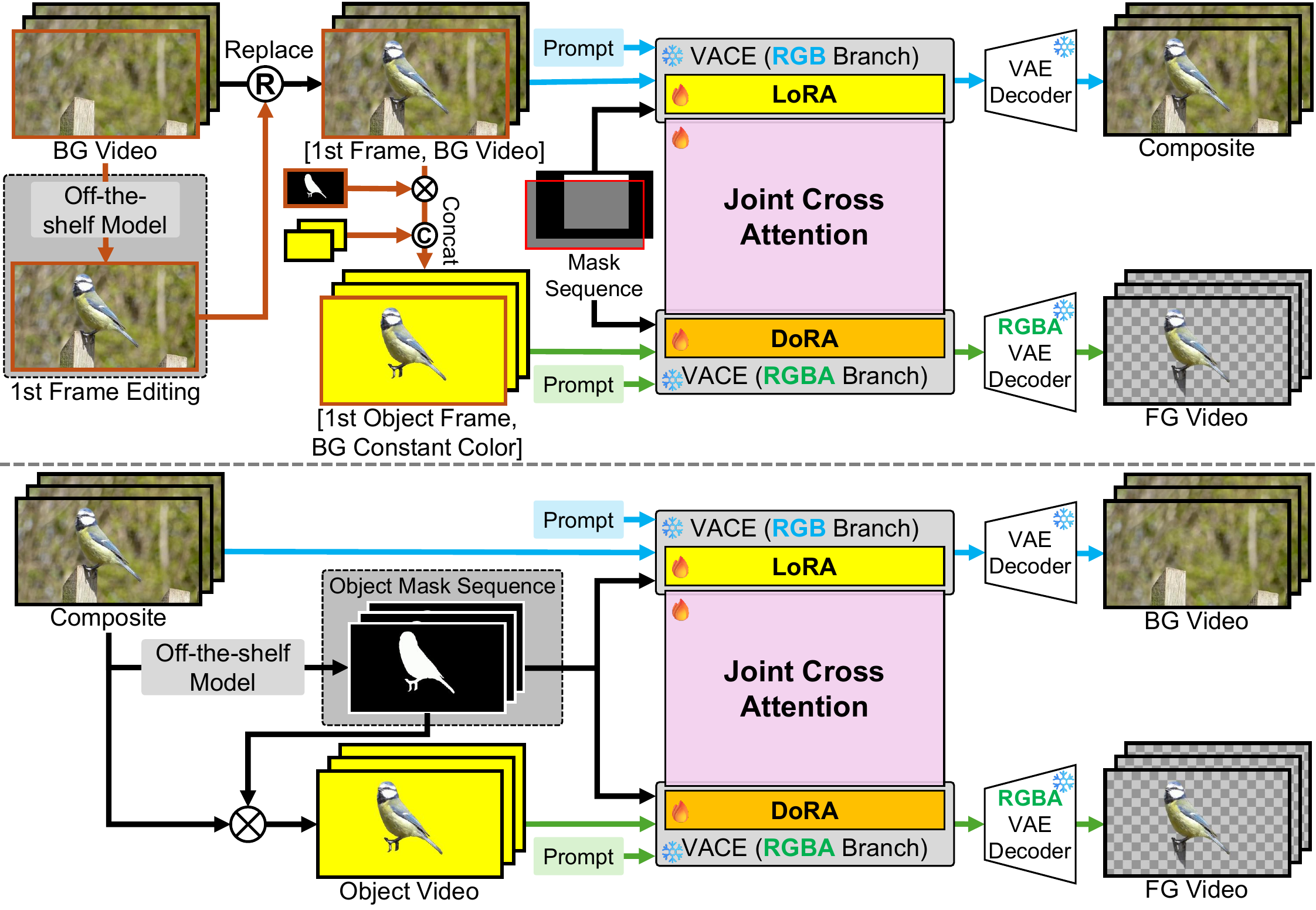}
    \vspace{-2.5ex}
    \caption{Overviews of \MethodNameInsert{} and \MethodNameDecompose{}. \emph{(Top) \MethodNameInsert{}} performs video layered object insertion by generating foreground layers alongside the composite. \emph{(Bottom) \MethodNameDecompose{}} performs video layer decomposition by separating the composite into foreground and background. 
    }
    \label{fig:model}
\end{figure*}

\section{\MethodName{} for Explicit Layer Modeling}

Our goal is to model videos in an explicit layered representation that separates the foreground object and its associated visual effects from the background scene. 
To this end, \MethodName{} adopts a dual-branch diffusion architecture with an RGB branch for scene-level appearance and an RGBA branch for foreground layer prediction. While the two branches operate on different layers, they are integrated through joint cross-attention \citep{li2024simple}, enabling bidirectional information exchange and consistent layered synthesis.
\MethodName{} provides a unified backbone for two layered video tasks: \MethodNameInsert{} for layered object insertion and \MethodNameDecompose{} for layered video decomposition. The two instantiations differ in their inputs, outputs, and training objectives as described below.

\subsection{\MethodNameInsert{}}
\MethodNameInsert{} applies the dual-branch architecture to layered object insertion, aiming to synthesize a foreground object layer $V_F$ that can be seamlessly integrated into a target background $V_B$.
The RGB branch produces an RGB composite $V_C$, while the RGBA branch generates an explicit foreground layer containing the object and its associated effects.
Here, $V_C$ provides a temporally coherent visualization and acts as an auxiliary signal during denoising, whereas $V_F$ serves as the definitive representation of the inserted object. 
At inference, $V_F$ is alpha-blended with $V_B$ to obtain the final high-fidelity result, making the method robust to minor background inconsistencies in $V_C$.

\MethodNameInsert{} takes four inputs: a background video $V_B$, a
text prompt $T$ describing the object, a sequence of bounding boxes $B$
indicating its location, and an edited first frame $E$ where the object has
already been inserted. $E$ can be synthesized by any modern text-guided image editing model~\citep{liu2025step1x, Nano-Banana-2},
and anchors the object’s appearance before temporal generation begins.
We first extract the foreground object from $E$ using SAM2~\citep{ravi2024sam}, obtaining an RGB image and its
binary alpha mask. These are used to build a conditioning video $C_{F}$ for the
RGBA branch: the first frame contains the extracted foreground over a
constant background color, and the remaining frames contain only the
constant background (with zero alpha). For the RGB branch, we replace
the first frame of the background video with $E$ to
form $C_C$. 
We also convert $B$ into a binary mask sequence $C_M$, with the first-frame mask set to zero to prevent modification.
The constructed conditioning signals are fed into the dual-branch model 
with text prompts. The RGB branch receives the original text prompt $T$, while the RGBA branch receives $T$ together with ``\texttt{The background is transparent.}'', encouraging it to generate an RGBA foreground layer.

To maintain consistency between layers, the branches are coupled through
joint cross-attention at every transformer block: the RGB branch attends
to RGBA features, and the RGBA branch symmetrically attends to RGB
features. This ensures that the composite reflects the geometry and
appearance of the foreground layer, while the foreground layer captures
scene-dependent cues such as lighting, motion, and occlusion. After
denoising, each branch’s latent representation is decoded by its
corresponding VAE, producing the final RGB composite and RGBA foreground
layer.

\subsection{\MethodNameDecompose{}}
\MethodNameDecompose{} instantiates our dual-branch diffusion framework
for layered video decomposition: given a composite video $V_C$, the model
recovers both a clean background video $V_B$ and an RGBA foreground layer $V_F$. The RGB branch removes the foreground object and its associated visual effects to reconstruct $V_B$, while the RGBA branch predicts
$V_F$, capturing appearance, boundaries, transparency, and effects such as shadows and reflections. Together, these outputs form an explicit layered representation of the input.

The model takes as input a composite video \(V_C\), a text prompt \(T\)
describing the foreground object, the object name \(T_N\), and an object
mask sequence \(M\). The masks can be obtained using off-the-shelf
video segmentation models such as SAM2~\citep{ravi2024sam} and need not
include object-induced effects. To guide the RGBA branch, we construct
an object-only conditioning video \(C'_{F}\) by masking out background
regions in \(V_C\), providing a coarse spatial prior while allowing the
model to refine appearance and transparency during denoising. The RGB
branch receives \(V_C\) as its conditioning video,
encouraging it to remove the object and restore occluded background
content. As in \MethodNameInsert{}, the object masks \(M\) are
provided to both branches. The RGBA branch receives the foreground
description \(T\), while the RGB branch receives the auxiliary
prompt ``\texttt{Remove the \{object\_name\} from the video naturally.}'' where \texttt{object\_name} is replaced with $T_N$.

\subsection{Training}


Both \MethodNameInsert{} and \MethodNameDecompose{} adopt the VACE~\citep{jiang2025vace} diffusion backbone for both branches, and the RGBA branch uses the RGBA VAE from WAN-alpha~\citep{dong2025wan}, which jointly encodes and decodes RGB and alpha.
Building on this architecture, we employ a hybrid LoRA--DoRA adaptation strategy to account for the distinct roles of the two branches. 
The RGB branch operates in-domain, predicting RGB videos within the pretrained latent space, and thus we adopt LoRA~\citep{hu2022lora} for efficient adaptation. 
In contrast, the RGBA branch must learn out-of-domain concepts such as transparency and alpha mattes.
To support this distribution shift, we adopt DoRA~\citep{liu2024dora}, which decouples weight magnitude and direction, leading to more stable optimization while preserving the pretrained directional prior.
Empirically, this hybrid strategy improves training stability and yields higher-fidelity layered representations across both tasks.
To further stabilize joint optimization, we train the model 
using disentangled timestep sampling~\citep{li2024simple}, 
which also enables conditional generation by independently controlling noise levels in each branch. 
For example, fixing the RGBA branch at low noise allows the model to edit foreground layers while preserving the background, enabling scene-aware layered manipulation (\cref{fig:application_figure}(b)).

\section{Experiments}

\begin{figure*}[t]
    \vspace{-5pt}
    \includegraphics[width=\textwidth]{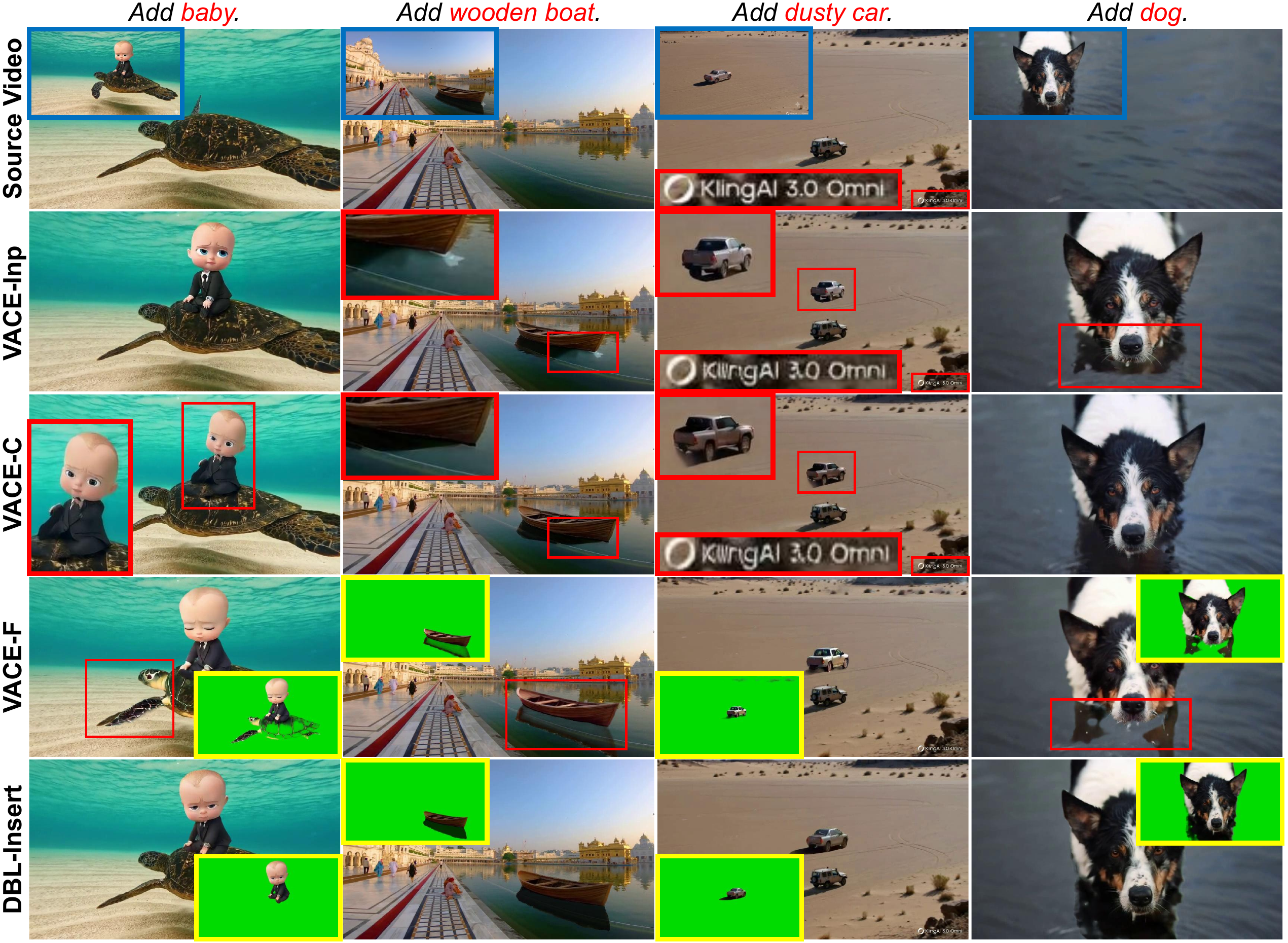}
    \vspace{-5ex}
    \caption{Qualitative video object insertion comparison with existing SOTA models. The blue and yellow boxes denote the edited first frame and foreground layer, respectively. Here, results of VACE-F and \MethodNameInsert{} are obtained by compositing the foreground layer onto the background video.
    }
    \label{fig:LayerVOI_Qual}
\end{figure*}

\subsection{\MethodNameInsert{}: Comparison and Ablation}
We evaluate layered object insertion using three categories of metrics:
(1) text–video alignment (ViCLIP-T~\citep{wang2022internvideo}),
(2) video quality (VBench~\citep{huang2025vbench++}) including
background/subject consistency, imaging quality, and aesthetics, and
(3) subject consistency (CLIP-I~\citep{radford2021learning}
and DINO-I~\citep{oquab2023dinov2}). We compare against
AnyV2V~\citep{kuanyv2v}, ReVideo~\citep{mou2024revideo}, and
VACE-Inpainting (VACE-Inp)~\citep{jiang2025vace}.
For evaluation of layered video tasks, we construct \emph{LayeredVid-Benchmark}, comprising 65 videos randomly sampled from TriLayer and held-out from training, covering diverse object categories and real-world scenarios.
Details and examples (Sec.~\ref{sec:benchmark} and \cref{fig:benchmark_examples}) are provided in the supplementary material.

To analyze the contribution of explicit triplet supervision, we include two VACE variants fine-tuned on \DatasetName{}: VACE-C (single-branch, background--composite supervision) and VACE-F (single-branch, background--foreground supervision).
Triplet supervision requires both a
\DatasetName{} dataset and a dual-branch architecture with joint
cross-attention, which together enable the RGB and RGBA branches to
learn disentangled scene and foreground representations. 
These components cannot be ablated independently: removing the RGBA branch or 
cross-attention collapses the model into a single-branch formulation. 
VACE-C and VACE-F therefore serve as controlled ablations 
with the same conditioning as \MethodNameInsert{} 
while removing triplet supervision and cross-branch interaction.

Quantitative and qualitative comparisons are shown in
\cref{tab:compare_voi_models,fig:LayerVOI_Qual}. AnyV2V and ReVideo
often alter background appearance or introduce temporal inconsistencies,
while VACE-Inp is limited by its inpainting formulation and cannot
synthesize object-induced effects outside the masked region. 
VACE-C and VACE-F benefit from dataset supervision but still underperform in subject and background consistency and compositing quality, exhibiting unstable visual effects and background appearance changes (e.g., text distortion and saturation shifts).
In contrast, \MethodNameInsert{} achieves the most balanced performance across all
metrics: the RGBA branch focuses on foreground layer synthesis, and the
RGB branch ensures scene-consistent compositing, producing more realistic
integration. Additional human preference study (\cref{tab:user_object_insertion}) and qualitative results (\cref{fig:more_qual_merge}(a) and~\ref{fig:more_various_insert}) are provided in the supplementary material.

\begin{table*}[t]
    \centering
    \caption{
    Video object insertion comparison on LayeredVid-Benchmark.
    }
    \resizebox{0.95\linewidth}{!}{
        \begin{tabular}{l|c|ccccc|cc}
        \toprule
        Model &
        ViCLIP-T $\uparrow$ &
        \makecell{Background\\Consistency $\uparrow$} &
        \makecell{Subject\\Consistency $\uparrow$} &
        \makecell{Motion\\Smoothness $\uparrow$} &
        Aesthetic $\uparrow$ &
        \makecell{Imaging\\Quality $\uparrow$} &
        CLIP-I $\uparrow$ &
        DINO-I $\uparrow$ \\
        \midrule
        AnyV2V~\citep{kuanyv2v}
        & 23.520 & 0.906 & 0.902 & 0.984 & 0.480 & 0.583 & 0.740 & 0.567 \\
        
        ReVideo~\citep{mou2024revideo}
        & 24.303 & 0.926 & 0.934 & 0.991 & 0.511 & \textbf{0.666} & 0.764 & 0.699 \\

        VACE-Inp~\citep{jiang2025vace}
        & 24.717 & 0.943 & 0.954 & 0.992 & 0.506 & 0.663 & 0.760 & 0.716 \\

        VACE-C
        & 24.596 & 0.944 & \textbf{0.963} & \textbf{0.993} & 0.522 & 0.661 & 0.781 & 0.739 \\

        VACE-F
        & 24.473 & 0.934 & 0.954 & 0.992 & 0.534 & 0.661 & 0.783 & 0.716 \\
        

        \cellcolor{blue!15}\MethodNameInsert{}
        & \cellcolor{blue!15}\textbf{24.794} & \cellcolor{blue!15}\textbf{0.948} & \cellcolor{blue!15}\textbf{0.963} & \cellcolor{blue!15}\textbf{0.993} & \cellcolor{blue!15}\textbf{0.535} & \cellcolor{blue!15}0.660 & \cellcolor{blue!15}\textbf{0.791} & \cellcolor{blue!15}\textbf{0.748} \\
        
        \bottomrule
    \end{tabular}
    }
    \label{tab:compare_voi_models}
\end{table*}
\begin{figure*}[t]
    \vspace{-5pt}
    \centering
    \includegraphics[width=\textwidth]{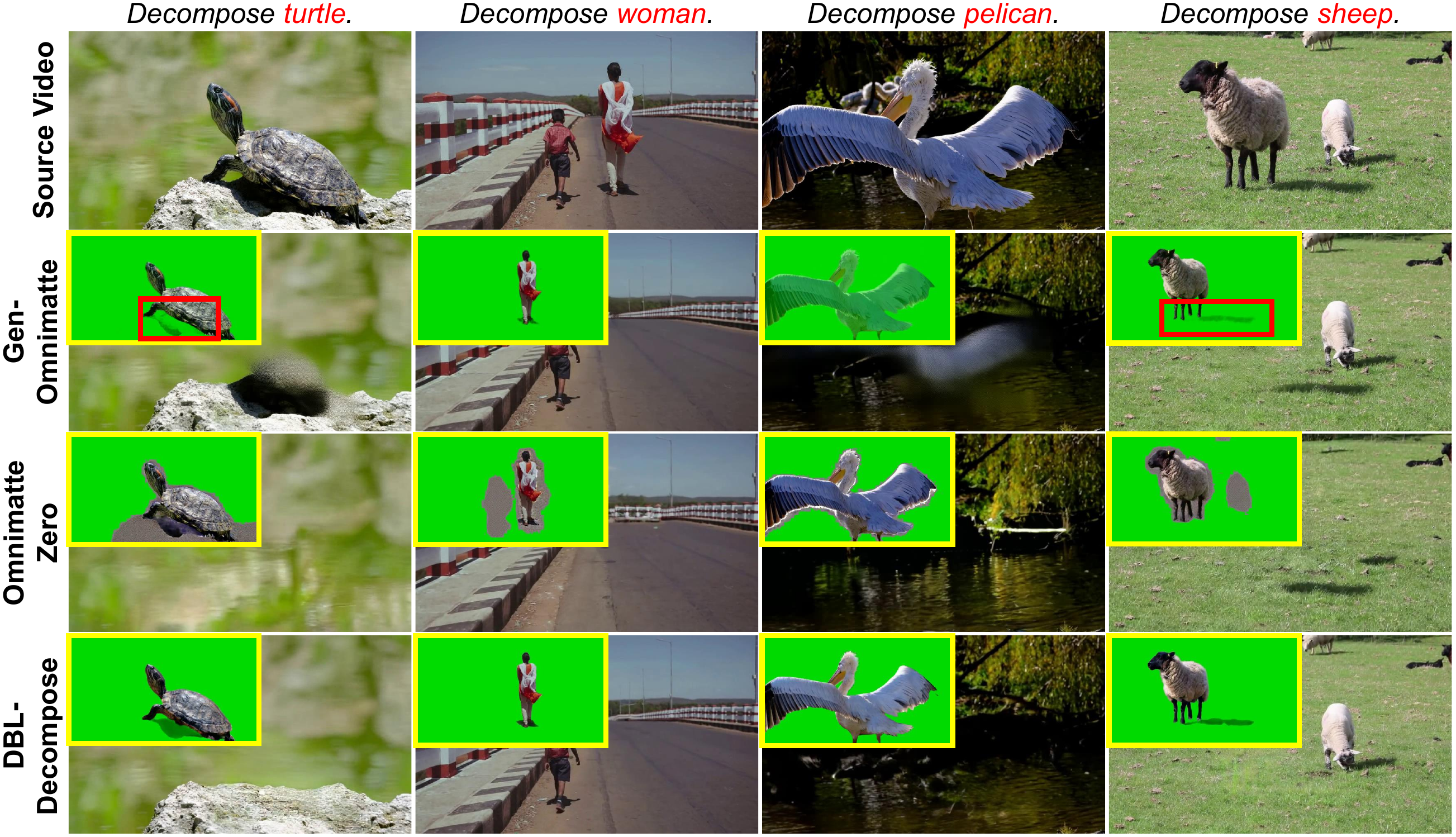}
    \vspace{-5ex}
    \caption{Qualitative video layer decomposition comparison with existing SOTA models. The yellow box denotes the foreground layer.
    }
    \label{fig:DBLDecomp_Qual}
\end{figure*}

\begin{table}[t]
    \centering
    \caption{Background layer reconstruction comparison on Movie and Kubric benchmarks.
    ``-" denotes missing values due to unreported data.
    }
    \small
    \resizebox{0.95\linewidth}{!}{
    \begin{tabular}{l|ccc|ccc}
        \toprule
        \multirow{2}{*}{Method}
        & \multicolumn{3}{c|}{Movie} & \multicolumn{3}{c}{Kubric} \\
        & 
        PSNR$\uparrow$ &
        LPIPS$\downarrow$ &
        SSIM$\uparrow$ &
        PSNR$\uparrow$ &
        LPIPS$\downarrow$ &
        SSIM$\uparrow$ \\
        \midrule

        \rowcolor{gray!10}
        \multicolumn{7}{l}{\textbf{Reconstruction-based \emph{omnimatte} methods}} \\

        Omnimatte~\citep{lu2021omnimatte} &
        21.76 & 0.239 & 0.736 &
        26.81 & 0.207 & 0.831 \\

        D2NeRF~\citep{wu2022d} &
        - & - & - &
        34.99 & 0.113 & 0.887 \\

        LNA~\citep{kasten2021layered} &
        23.10 & 0.129 & 0.847 &
        - & - & - \\

        OmnimatteRF~\citep{lin2023omnimatterf} &
        33.86 & 0.017 & 0.981 &
        40.91 & 0.028 & 0.970 \\

        Generative Omnimatte~\citep{lee2025generative} &
        32.69 & 0.030 & 0.989 &
        44.07 & 0.010 & 0.981 \\

        OmnimatteZero~\citep{samuel2025omnimattezero} &
        35.11 & 0.014 & 0.992 &
        44.97 & 0.010 & 0.988 \\

        \midrule

        \rowcolor{gray!10}
        \multicolumn{7}{l}{\textbf{Generation-based \emph{diffusion} methods}} \\

        ObjectDrop~\citep{winter2024objectdrop} & 
        28.05 & 0.124 & - &
        34.22 & 0.083 & - \\

        Lumiere inpainting~\citep{bar2024lumiere} & 
        26.62 & 0.148 & - &
        31.46 & 0.157 & - \\

        Propainter~\citep{zhou2023propainter} & 
        27.44 & 0.114 & - &
        34.67 & 0.056 & - \\

        DiffuEraser~\citep{li2025diffueraserdiffusionmodelvideo} & 
        29.51 & 0.105 & - &
        35.19 & 0.048 & - \\

        \rowcolor{blue!15}
        \MethodNameDecompose{} &
        \textbf{33.42} & \textbf{0.017} & \textbf{0.971} &
        \textbf{38.78} & \textbf{0.021} & \textbf{0.977} \\

        \bottomrule
    \end{tabular}
    }
    \label{tab:omnimatte_comparison}
\end{table}

\begin{figure*}[t]
    \vspace{-5pt}
    \centering
    \includegraphics[width=0.95\textwidth]{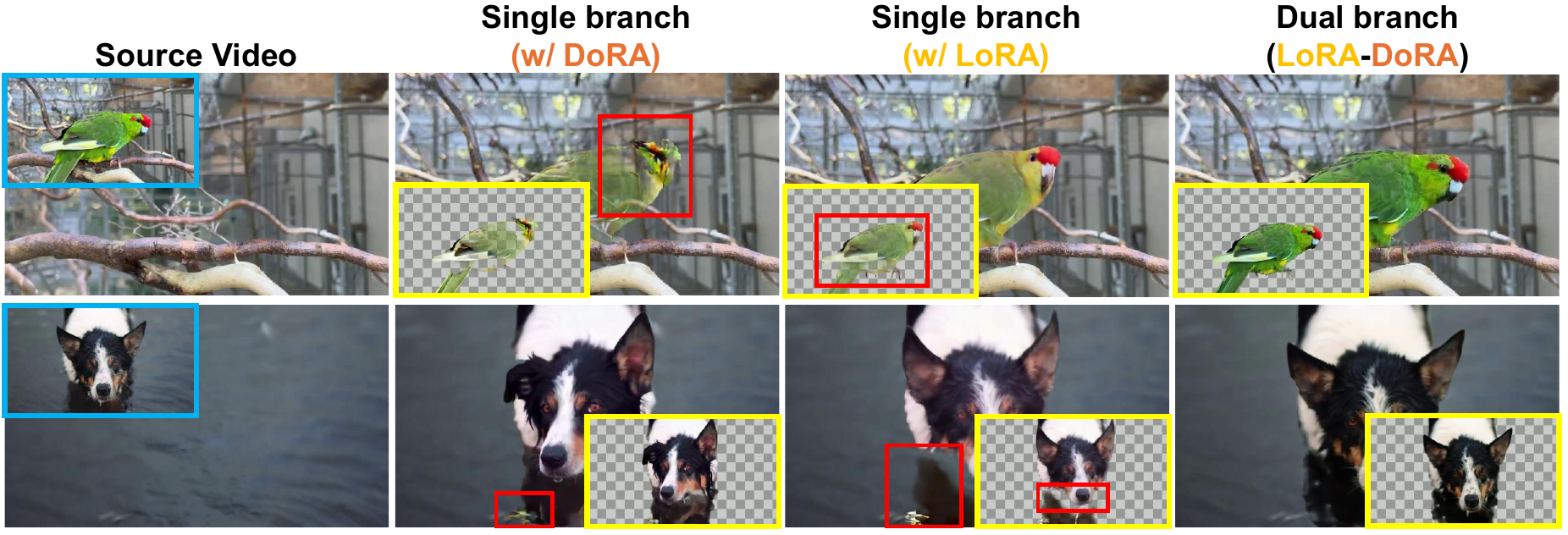}
    \vspace{-3ex}
    \caption{Qualitative ablation on the effect of dual-branch modeling. The blue and yellow boxes denote the first frame and foreground layer, respectively.}
    \label{fig:abl_single_branch}
\end{figure*}
\begin{table}[t]
    \centering
    \caption{LoRA-DoRA ablation on video layered object insertion.
    }
    \resizebox{0.95\linewidth}{!}{
        \begin{tabular}{l|c|ccccc|cc}
        \toprule
        Method (RGB-RGBA) &
        ViCLIP-T $\uparrow$ &
        \makecell{Background\\Consistency $\uparrow$} &
        \makecell{Subject\\Consistency $\uparrow$} &
        \makecell{Motion\\Smoothness $\uparrow$} &
        Aesthetic $\uparrow$ &
        \makecell{Imaging\\Quality $\uparrow$} &
        CLIP-I $\uparrow$ &
        DINO-I $\uparrow$ \\
        \midrule
        
        DoRA-DoRA
        & 24.520 & 0.927 & 0.935 & 0.993 & 0.525 & 0.641 & 0.774 & 0.672 \\

        LoRA-LoRA
        & 24.681 & 0.938 & 0.954 & \textbf{0.994} & \textbf{0.539} & 0.657 & 0.775 & 0.742 \\

        DoRA-LoRA
        & 24.604 & 0.935 & 0.944 & 0.993 & \textbf{0.539} & \textbf{0.664} & 0.788 & 0.701 \\

        \cellcolor{blue!15}\MethodNameInsert{} (LoRA-DoRA)
        & \cellcolor{blue!15}\textbf{24.794} & \cellcolor{blue!15}\textbf{0.948} & \cellcolor{blue!15}\textbf{0.963} & \cellcolor{blue!15}0.993 & \cellcolor{blue!15}0.535 & \cellcolor{blue!15}0.660 & \cellcolor{blue!15}\textbf{0.791} & \cellcolor{blue!15}\textbf{0.748} \\

        \bottomrule
        \end{tabular}
    }
    \label{tab:lora_dora_ablation}
\end{table}

\begin{figure*}[t]
    \vspace{-3pt}
    \centering
    \includegraphics[width=\textwidth]{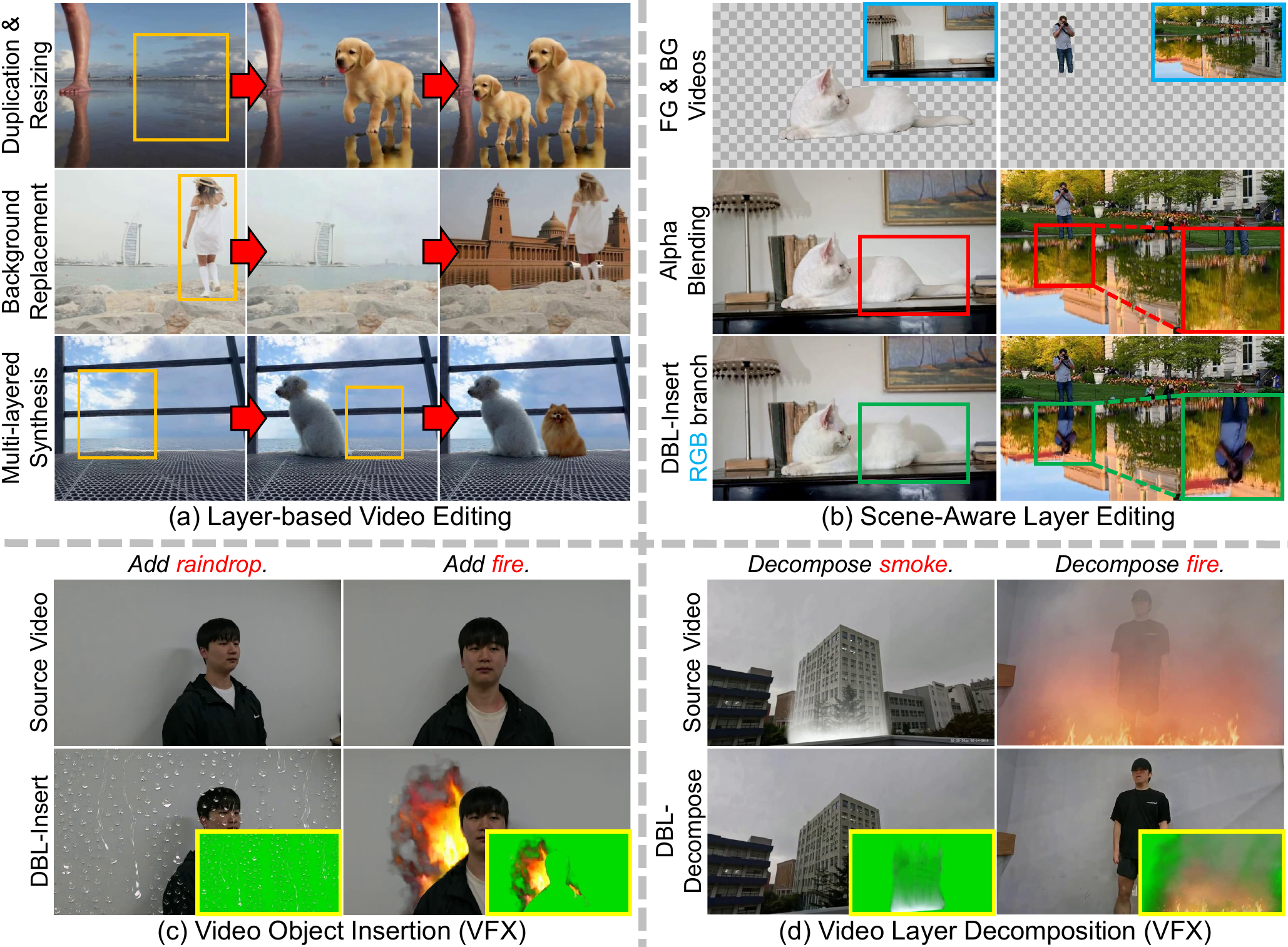}
    \vspace{-4.5ex}
    \caption{Qualitative application examples. (a) Layer-based video editing. (b) Scene-aware layer editing. (c) and (d) VFX insertion and decomposition, respectively. The orange, blue, and yellow boxes denote the input mask, background layer, and foreground layer, respectively.
    }
    \label{fig:application_figure}
\end{figure*} 

\subsection{\MethodNameDecompose{}: Foreground and Background Layer Reconstruction}

We evaluate layer decomposition using PSNR, SSIM, and LPIPS for background and foreground reconstruction. 
For background reconstruction, we follow prior work and evaluate on the
Movies~\citep{lin2023omnimatterf} and Kubric~\citep{wu2022d} benchmarks,
comparing against generation-based and reconstruction-based methods.
For foreground reconstruction, we compare against two representative state-of-the-art methods, 
Gen-Omnimatte~\citep{lee2025generative} and OmnimatteZero~\citep{samuel2025omnimattezero}, which reflect optimization-based and feed-forward paradigms, respectively.

Background reconstruction results are summarized in \cref{tab:omnimatte_comparison}. Among generation-based methods, \MethodNameDecompose{} achieves the best performance across all metrics. While reconstruction-oriented methods such as Gen-Omnimatte and OmnimatteZero achieve higher PSNR/SSIM/LPIPS scores, this is expected since they are explicitly designed to reconstruct the input video by optimizing pixel-level fidelity or preserving latent representations through attention masking.
In contrast, \MethodNameDecompose{} leverages diffusion priors to generate perceptually plausible and visually coherent background layers rather than directly reproducing the ground-truth frames.
As shown in \cref{fig:DBLDecomp_Qual}, our method produces cleaner and more realistic backgrounds while maintaining high-quality foreground layers with accurate object boundaries and scene-dependent effects. 
However, Gen-Omnimatte and OmnimatteZero often produce semi-transparent or incomplete foreground layers, particularly around complex visual effects and fine-scale boundaries.
These results demonstrate that \MethodNameDecompose{} provides a more effective decomposition of both foreground and background layers, despite the inherent trade-off in pixel-wise reconstruction metrics.
Additional perceptual evaluations, including Fréchet Inception Distance (FID) and a human preference study (\cref{tab:fid_comparison} and~\ref{tab:user_layer_decomposition}), as well as qualitative results (\cref{fig:more_LD_comparison} and~\ref{fig:more_qual_merge}(b)), are provided in the supplementary material.

\subsection{Additional Ablation Studies and Applications}

\paragraph{Ablation: LoRA vs. DoRA.}
We compare different adapter configurations for the RGB and RGBA branches (\cref{tab:lora_dora_ablation}).
LoRA effectively preserves pretrained in-domain representations, making it suitable for the RGB branch. 
In contrast, DoRA's decoupled weight parameterization provides stronger adaptation capability for the RGBA branch, which undergoes a larger domain shift due to alpha-layer modeling. 
The proposed LoRA-DoRA configuration achieves the most balanced performance, particularly improving semantic text-alignment and background/subject consistency. 
These results demonstrate that branch-specific adapter selection is important for effective video layered object insertion.
Additional qualitative ablations (\cref{fig:lora_dora}) are provided in the supplementary material.

\vspace{-2pt}
\paragraph{Ablation: Dual-branch structure.}
We further compare the single- and dual-branch structures for video layered object insertion (\cref{fig:abl_single_branch}). In the single-branch structure, RGB and RGBA latents are concatenated and jointly denoised, then decoded by their respective VAEs.
The single-branch structure often produces inconsistent object appearance, semi-transparent regions, unstable visual effects, and noticeable artifacts, as jointly modeling RGB and RGBA latents makes it difficult to sufficiently learn their distinct representations.
In contrast, our dual-branch structure explicitly separates the two representations while enabling information exchange through joint cross-attention, resulting in higher-quality and more consistent foreground layers. Additional quantitative ablation (\cref{tab:dual_branch}) is provided in the supplementary material.

\vspace{-2pt}
\paragraph{Ablation: Background refinement.}
We investigate the effect of using a refined background video dataset during training (\cref{fig:more_agbi_effect,tab:background_refinement_ablation}). We compare models trained with and without applying \MethodNameRefinement{} to the backgrounds generated by the object removal model. Without the refined dataset, the trained model often produces backgrounds with residual traces of removed objects and unnatural artifacts.
In contrast, training with the refined dataset produces cleaner and more natural backgrounds with substantially fewer residual artifacts. 
These results demonstrate that \MethodNameRefinement{} produces higher-quality background training data and contributes to improved background reconstruction quality. 

\vspace{-2pt}
\paragraph{Application: Layer-based and scene-aware editing.}
Our layered representation enables intuitive video editing by allowing users to independently stylize the background or foreground while preserving the other layer (\cref{fig:application_figure}(a)). These edits remain temporally consistent and can be freely recomposed without manual matting or complex post-processing. Furthermore, repeated application of our framework enables selective editing and coherent recomposition of individual scene elements.
In addition, \MethodNameInsert{} supports scene-aware layer editing, automatically synthesizing scene-consistent visual effects conditioned on the given foreground layer (\cref{fig:application_figure}(b)). This allows the model to generate object-dependent effects, such as shadows, reflections, and light interactions, enabling inserted or modified objects to blend naturally into the scene with high fidelity and without manual design.

\vspace{-2pt}
\paragraph{Application: VFX editing and decomposition.}
\MethodName{} is not limited to solid foreground objects, but extends to
a broader range of foreground phenomena, including semi‑transparent VFX
elements such as fire, raindrops, and smoke. To validate this capability, we train
both \MethodNameInsert{} and \MethodNameDecompose{} on an internal
dataset of 1,155 triplet video clips, consisting of real‑world
background videos paired with rendered VFX elements and related effects, 
while reserving 25 videos for testing. 
\cref{fig:application_figure}(c--d) presents examples of
layered object insertion and layer decomposition, demonstrating that our
framework can effectively handle VFX‑style semi‑transparent phenomena as well.
\vspace{-15pt}
\section{Conclusion}
\vspace{-3.5pt}
In this work, we introduce a novel triplet video dataset \DatasetName{} that explicitly models foreground–background–composite relationships, enabling supervision for layered video representations. Building upon this dataset, we propose two models: \MethodNameInsert{} for video layered object insertion and \MethodNameDecompose{} for video layer decomposition. Our approach achieves strong performance across multiple metrics while supporting practical and flexible layer-based video editing. We believe that our dataset provides a valuable resource for future research on layer-aware video editing.
Despite these promising results, our models still struggle to capture complex physical interactions and highly dynamic motions between objects and scenes, and the dual-branch architecture introduces additional computational and memory overhead. Future work will focus on improving efficiency and better modeling complex physical interactions.

\clearpage
\clearpage
\section*{AI Disclosure} 
We used generative AI tools to assist with and polish the writing of this manuscript, including improving clarity, grammar, and readability. All AI-assisted text was reviewed and revised by the authors, who take full responsibility for the final content of the paper.

\bibliography{main}
\bibliographystyle{iclr2027_conference}

\clearpage
\appendix

\renewcommand{\thefigure}{S\arabic{figure}}
\renewcommand{\thetable}{S\arabic{table}}
\renewcommand{\theequation}{S\arabic{equation}}
\setcounter{figure}{0}
\setcounter{table}{0}
\setcounter{equation}{0}

\begin{center}
{\LARGE \bfseries Supplementary Material}\\
\end{center}
\vspace{1em}


\section{Related Work}

\paragraph{Image layered editing.}
Existing image-layered generation methods have explored incorporating transparency and layered representations into diffusion-based generative models.
LayerDiffuse~\citep{zhang2024transparent} introduces \emph{latent transparency}, which encodes the alpha channel as a latent offset while preserving the latent distribution of a pretrained latent diffusion model. 
Qwen-Image-Layered~\citep{yin2026qwen} further employs an RGBA-VAE to encode RGB images and RGBA layers into a shared latent space, reducing the distribution gap between them, and adopts an MMDiT-based architecture with layer-aware positional representations for flexible multi-layer generation.
LaDe~\citep{lungu2026lade} extends this direction to unified multi-layer media generation and decomposition using a 4D RoPE positional encoding that links the content information to its respective layer.
LASAGNA~\citep{yang2026controllable} focuses on controllable layered image editing by jointly generating composite images, backgrounds, and transparent foregrounds with visual effects such as shadows and reflections under diverse conditioning inputs.
Despite these advances, layered video generation remains largely unexplored.
One key challenge lies in the lack of video datasets with RGBA layer annotations, which are substantially more difficult to obtain than static layered images.
To address these challenges, we construct a video-layered triplet dataset consisting of composite, background, and RGBA foreground videos, providing consistent layer-level supervision for learning coherent video layer decomposition and layered object insertion.

\paragraph{Video object insertion.}
Existing works approach video object insertion through several paradigms.
Training-free methods~\citep{kuanyv2v, li2025flowdirector} inject
object-related features into pretrained diffusion models, enabling
insertion without additional training. Inpainting-based approaches
formulate insertion as masked video completion~\citep{bian2025videopainter},
while motion-guided methods~\citep{mou2024revideo, tu2025videoanydoor}
use object trajectories to improve temporal coherence and controllability.
Recent systems further expand applicability: OmniInsert~\citep{chen2025omniinsert}
supports mask-free insertion of flexible objects, InsertAnywhere~\citep{jin2025insertanywhere}
estimates object masks via 4D reconstruction for static objects, and
LoVoRA~\citep{xiao2025lovora} leverages optical flow to guide insertion.
Despite these advances, existing methods operate purely in the RGB
domain and do not explicitly model object layers, limiting layer-aware
editing and hindering realistic compositing in complex scenes. In
contrast, our approach jointly models object insertion and layered video
representations within a unified diffusion framework.


\paragraph{Video layer decomposition.}
Video layer decomposition aims to separate composite videos into
foreground and background layers for applications such as object
removal, duplication, and background replacement.
Omnimatte~\citep{lu2021omnimatte} introduced a framework for extracting
foreground layers that include both object appearance and associated
visual effects. OmnimatteRF~\citep{lin2023omnimatterf} incorporates a
neural radiance field for improved consistency, and
Omnimatte3D~\citep{suhail2023omnimatte3d} extends this idea to 3D scene
representations. 
Recently, Gen-Omnimatte~\citep{lee2025generative} performs per-scene optimization using composite--background pairs to obtain foreground layers, 
while OmnimatteZero~\citep{samuel2025omnimattezero} is training-free and leverages a pretrained video diffusion model with self-attention masking to isolate foreground objects and their associated effects.
More recently, EasyOmnimatte~\citep{hu2026easyomnimatte} introduces alpha-matte supervision and fine-tunes a pretrained video inpainting model to extract an alpha-matte, while relying on the pretrained inpainting model for background reconstruction.
However, these approaches do not directly learn aligned composite--foreground--background triplet supervision, leaving the decomposition problem under-constrained and leading to unstable decomposition and limited generalization. 
In contrast, our approach introduces explicit triplet supervision through \DatasetName{}, enabling
generalizable learning of layered video representations.

\section{Details of Implementation}

\paragraph{Implementation.}
We adopt VACE-1.3B~\citep{jiang2025vace} as our base diffusion transformer. Both \MethodNameInsert{} and \MethodNameDecompose{} are
trained for 3 epochs using AdamW~\citep{loshchilov2017decoupled} with a
learning rate of $1\times10^{-4}$. LoRA~\citep{hu2022lora} (rank 128) is
applied to the RGB branch, while DoRA~\citep{liu2024dora} (rank 32) is
applied to the RGBA branch. We set the flow shift to 3.0 in training and inference.
For VFX-specific applications, we further fine-tune the models pretrained on TriLayer using an internal dataset of layered VFX elements. We train the model on 4 NVIDIA A100 80GB GPUs with a total batch size of 8. 

\paragraph{Joint cross attention.} 
The joint cross-attention weights are initialized from the self-attention weights of the pretrained base model (VACE)~\citep{jiang2025vace}. 
During training, we apply LoRA to the query, key, and value projections of the RGB branch (i.e., $W_Q^x, W_K^x, W_V^x$), and DoRA to those of the RGBA branch (i.e., $W_Q^y, W_K^y, W_V^y$).

\paragraph{RGBA branch conditioning.}
Following the RGBA VAE~\citep{dong2025wan}, we use a fixed set of eight constant background colors for the RGBA branch: \texttt{\{black, blue, green, cyan, red, magenta, yellow, white\}}. We randomly select the background color during both training and inference.

\paragraph{VACE-C and VACE-F training.}
VACE-C denotes VACE fine-tuned on the \DatasetName{} dataset for video object insertion, and VACE-F denotes VACE fine-tuned on the \DatasetName{} dataset for video layered object insertion.
For VACE-C, we fine-tune the diffusion transformer using LoRA, while for VACE-F, we adopt DoRA for fine-tuning. For fair comparison, both models are trained on the same \DatasetName{} and internal datasets with identical training settings.

\paragraph{Compared methods.} For AnyV2V~\citep{kuanyv2v}, we concatenate the first frame with the background video as the input. ReVideo~\citep{mou2024revideo} supports generating up to 14 frames at a time. To produce an 81-frame video, we therefore adopt an auto-regressive strategy, in which the last frame of the previously generated clip is used as the first frame for the next generation step. 

Comparisons with OmniInsert~\citep{chen2025omniinsert}, InsertAnywhere~\citep{jin2025insertanywhere}, LoVoRA~\citep{xiao2025lovora}, VideoAnydoor~\citep{tu2025videoanydoor}, and EasyOmnimatte~\citep{hu2026easyomnimatte} are infeasible because no official code is publicly available.

\paragraph{Evaluation.} 
ViCLIP-T~\citep{wang2022internvideo}, CLIP-I~\citep{radford2021learning}, and DINO-I~\citep{oquab2023dinov2} are evaluated using pretrained models. Specifically, we adopt \emph{ViCLIP-L\_InternVid-FLT-10M} for ViCLIP-T, while \emph{clip-vit-large-patch14} and \emph{dinov2-large} are used for CLIP-I and DINO-I, respectively. For CLIP-I and DINO-I, similarity is computed between the segmented object in the first frame and subsequent frames. For comparisons of foreground layer reconstruction, we provide the ground-truth background video and object masks as inputs to Gen-Omnimatte~\citep{lee2025generative} and OmnimatteZero~\citep{samuel2025omnimattezero}.

\paragraph{Inference.} 
During inference, we use separate negative prompts for the RGB and RGBA branches. The RGB branch uses \texttt{"Bright tones, overexposed, static, blurred details, subtitles, style, works, paintings, images, static, overall gray, worst quality, low quality, JPEG compression residue, ugly, incomplete, extra fingers, poorly drawn hands, poorly drawn faces, deformed, disfigured, misshapen limbs, fused fingers, still picture, messy background, three legs, many people in the background, walking backwards, changing color"}, while the RGBA branch uses \texttt{"Bright tones, overexposed, static, blurred details, subtitles, style, works, paintings, images, static, overall gray, worst quality, low quality, JPEG compression residue, ugly, incomplete, extra fingers, poorly drawn hands, poorly drawn faces, deformed, disfigured, misshapen limbs, fused fingers, still picture, messy background, three legs, many people in the background, walking backwards, changing color, body transparency"}. 
During inference, we use 50 sampling steps and set the classifier-free guidance (CFG) scale as 5.0. All generated videos contain 81 frames at a resolution of $480\times832$. 
Due to the dual-branch diffusion transformer architecture, our model requires approximately one hour per inference case. All inference experiments were conducted on a single NVIDIA A100 80GB GPU.

\begin{figure*}[t]
\centering
\begin{tcolorbox}[width=\textwidth]
You are a video analysis and evaluation assistant specialized in selecting high-quality videos suitable for object insertion tasks. Your job is to strictly assess each video based on the video frames, and determine whether the video is appropriate for training or evaluating object insertion models.

Please follow the detailed criteria below to make your judgment:

Rejection Criteria – A video is NOT suitable if:

1. The target object is completely absent from the video (i.e., it does not appear in any frame).

2. The object is too small or too large to be clearly visible or identifiable.

3. There are too many instances of the same object, making insertion ambiguous or cluttered.

4. (Important) The camera motion is not static and changes dynamically, resulting in significant background changes (e.g., 
switching from underwater to land, or from sky to ground).

5. The video contains blur or dynamic motion in the camera or the object, which makes the object or background unclear or unstable.

6. The overall video quality is low (e.g., object disappears, hard to see, excessive motion, or rapid frame transitions).

7. There is significant occlusion of the object or background in the video, making insertion unsuitable.

8. The object contains complex, dynamically changing textures (e.g., fluttering fur, layers of feathers), making realistic object insertion unreliable.

9. The object is not clearly visible or identifiable in the video, due to issues like distance, lighting, resolution, or visual ambiguity.

10. (Important) The object completely disappears from the visible frame during the video.

11. The background changes rapidly or inconsistently, causing disruption in scene understanding.

Examples English Output:

1. REJECTED: The object is not present in some frames of the video.

2. REJECTED: The object is too small or too large and partially occluded in the video.

3. REJECTED: The background changes rapidly in the video.

4. REJECTED: The object completely disappears from the visible frame during the video.

5. ACCEPTABLE: The object appears clearly in the video with sufficient size and no occlusion.

You must always output in English and strictly evaluate based on the above criteria.

\blue{\{Given video frames (user)\}}

Evaluate whether the video is suitable for an Object Insertion task according to the criteria and provide the ACCEPTABLE or REJECTED status along with the reason, as shown in the examples.
\end{tcolorbox}
\vspace{-3ex}
\caption{Instruction used for VLM-based evaluator of overall video and object quality. \blue{Blue text} indicates user input.}
\label{tab:prompt_filtering}
\end{figure*}
\begin{figure*}[t]
\centering
\begin{tcolorbox}[width=\textwidth]
You are an assistant that determines whether the major objects described in the given description is mentioned in the video title. If any of the key objects mentioned in the description match the video title, output 'YES'; if not, output 'NO'.

In either case, provide a brief explanation.

The output format is:

{YES/NO}: {Brief explanation}

Given description: \red{\{Description of the VLM-based evaluator for given video. (VLM)\}}

Video title: \blue{\{Given video title (user)\}}
\end{tcolorbox}
\vspace{-3ex}
\caption{Instruction used for VLM-based evaluator of major object quality. \blue{Blue text} indicates user input, and \red{red text} indicates VLM output of previous stage.}
\label{tab:prompt_object_filtering}
\end{figure*}
\begin{figure*}[t]
\centering
\begin{tcolorbox}[width=\textwidth]
You are an assistant that identifies the single most clearly visible main object from a given images and its title. Your task is to output only one object that is most prominently and clearly visible in the image. If no identifiable object is visible, return "Nothing".

Rules:

1. Only output one word or phrase: the most clearly visible object.

2. Ignore background elements unless they are the only visible content.

3. If multiple objects are present, choose the most salient or centrally focused one.

4. If the image is unclear, blurred, or contains no recognizable object, output "Nothing".

Examples English Output:

1. Object: ["dog"]

2. Object: ["human"]

3. Object: ["horse"]

4. Object: ["Nothing"]

You must always output in English and strictly evaluate based on the above criteria.

\blue{\{Given video frames (user)\}}

Video title: \blue{\{Given video title (user)\}}
\end{tcolorbox}
\vspace{-3ex}
\caption{Instruction used for VLM-based selector of major object in given video. \blue{Blue text} indicates user input.}
\label{tab:prompt_object_select}
\end{figure*}
\begin{figure*}[t]
\centering
\begin{tcolorbox}[width=\textwidth]
You are a prompt optimization expert. Your goal is to generate a high-quality and richly detailed English prompt that clearly describes the motion and visual characteristics of the main object based on the user-provided video information. Use the object and scene details to rewrite the prompt in a vivid, cinematic, or stylistically accurate way. You must strictly follow the formatting and tone of the examples below.

Task Requirements:

1. Clearly describe the motion or actions of the main object in the video using simple, direct verbs.

2. Enrich the prompt with key subject characteristics – including appearance, emotion, quantity, ethnicity, posture, etc.

3. Emphasize any camera movements or angles, such as tracking shots, aerial views, pans, or zoom-ins.

4. Accurately portray natural motion and spatial composition relevant to the subject (e.g., a dog running on grass, a bird soaring through sky).

5. Integrate visible details from the video such as clothing, objects, scenery, lighting, and background composition.

6. The written prompt should be about 80–100 words in length.

7. No matter what language the user provides as input, your rewritten prompt must always be in English.

Example of the written English prompt:

1. The video features a red sports car in a studio setting. The car is sleek and modern, with a low and wide stance. It has a large rear wing and a prominent front splitter. The wheels are black with a multi-spoke design. The car is positioned at a slight angle to the camera, allowing a view of both the front and side. The background is a neutral gray, which contrasts with the car's vibrant red color. The lighting is soft and even, highlighting the car's curves and contours. The overall style of the video is clean and professional, with a focus on the car's design and features.

2. The video shows a small black and brown dog with large ears and a green collar. The dog is resting its head on a person's arm, which is wearing a brown leather watch. The dog's eyes are wide open and it appears to be looking directly at the camera. The background is blurred but it seems to be an indoor setting with a white surface. The style of the video is a close-up shot with a shallow depth of field, focusing on the dog's face and the person's arm. The dog's expression is curious and attentive.

3. In the video, a man and a woman are seen walking together in a parking lot. The man is wearing a blue shirt and is carrying a pink bag, while the woman is dressed in a white shirt and sunglasses. They are walking towards a silver car, which is parked in the background. The parking lot is filled with other cars, indicating a busy area. The couple appears to be in a good mood, as they are smiling and enjoying each other's company. The overall style of the video is casual and candid, capturing a moment of everyday life.

Directly output the written English text in 80-100 words.

\blue{\{Given video frames (user)\}}

The main object is \red{\{object name\}}. Please write the high-quality and richly detailed English prompt.
\end{tcolorbox}
\vspace{-3ex}
\caption{Instruction used for VLM-based prompter of given video and object. \blue{Blue text} indicates user input, and \red{red text} indicates VLM output of previous stage.}
\label{tab:prompt_maker}
\end{figure*}
\begin{figure*}[t]
\centering
\begin{tcolorbox}[width=\textwidth]
You are a prompt optimization expert. Your goal is to generate a realistic and descriptive English prompt that accurately describes the overall visual scene of a user-provided video, with particular emphasis on the background, surrounding environment, and major objects. The generated prompt will be used as a textual description for video inpainting, so it should provide clear and useful information about the scene and background represented in the video.

Task Requirements:

1. Describe the overall situation and visual composition of the video in a realistic and natural manner.

2. Clearly describe the main background and environment of the video, including important landscapes, buildings, vegetation, roads, sky, water, indoor spaces, or other major scene elements.

3. Describe the main and surrounding objects that are visually important, including their position, state, and natural actions or movements when relevant.

4. Capture meaningful temporal motion in the video, such as walking people, moving animals, flowing water, swaying plants, or other naturally changing elements.

5. Prioritize observable scene and background information. Do not hallucinate hidden content or describe information that is not visually supported by the video.

6. Avoid overly detailed descriptions of the main foreground object when they are not useful for understanding the surrounding scene.

7. Write a coherent, concise prompt of approximately 50–80 words. Always output the prompt in English, regardless of the input language, and directly provide only the generated prompt without explanations.

Example of the written English prompt:

1. A peaceful grassy meadow stretches across the scene, surrounded by trees and low hills in the distance. A small lamb grazes quietly, lowering its head and slowly moving through the lush green grass. The lamb continues grazing as it gradually moves across the meadow, while the surrounding landscape remains consistent throughout the scene.

2. A sandy beach extends toward a calm blue ocean, with a rocky coastline visible in the distance. A dog sits near the shoreline, facing the water as gentle waves roll onto the sand. The dog remains seated and occasionally moves its head while the waves continuously move along the shore.

3. A wide river flows through a green valley, with rocky banks and dense vegetation along both sides. Several large trees grow near the water, while distant hills rise behind the river. The water continuously moves downstream, creating a natural flow through the center of the scene.

Directly output the written English text.

\blue{\{Given video frames (user)\}}

Describe the provided video according to the instructions.
\end{tcolorbox}
\vspace{-3ex}
\caption{Instruction used by the VLM-based prompter to generate a background refinement prompt from a given object-removed background video. \blue{Blue text} indicates the user input.}
\label{tab:prompt_inp_maker}
\end{figure*}

\section{Details of \DatasetName{} Dataset}
\label{sec:dataset}

\paragraph{VLM-based filtering.} We employ Qwen2.5-VL-32B-Instruct~\citep{bai2025qwen25vltechnicalreport} as the vision-language model (VLM) for data filtering. The filtering pipeline consists of two stages: (1) quality assessment and (2) object selection. In the quality assessment stage, as shown in \cref{tab:prompt_filtering}, the VLM evaluates both video and object quality by considering factors such as visibility, blurry, occlusion, and object prominence. Based on the generated reasoning, each sample is classified as \emph{ACCEPTABLE} or \emph{REJECTED}, and only accepted samples are retained. However, we observe that the VLM occasionally focuses on objects that are not visually salient or not properly captured by the camera, leading to unreliable reasoning. To address this issue, we introduce an object filtering stage. As shown in \cref{tab:prompt_object_filtering}, we identify the major object by comparing the VLM-generated reasoning with the video title, allowing us to filter out unsuitable object candidates. Finally, to assign a consistent object class label for each video, we employ a VLM-based selector, as shown in \cref{tab:prompt_object_select}, which determines the most representative object category for the given video.

\paragraph{VLM-based prompting.} As shown in \cref{tab:prompt_maker}, we design object-centric prompts using a VLM to focus on the objects present in the scene.

\begin{figure*}[t]
    \centering
    \includegraphics[width=\textwidth]{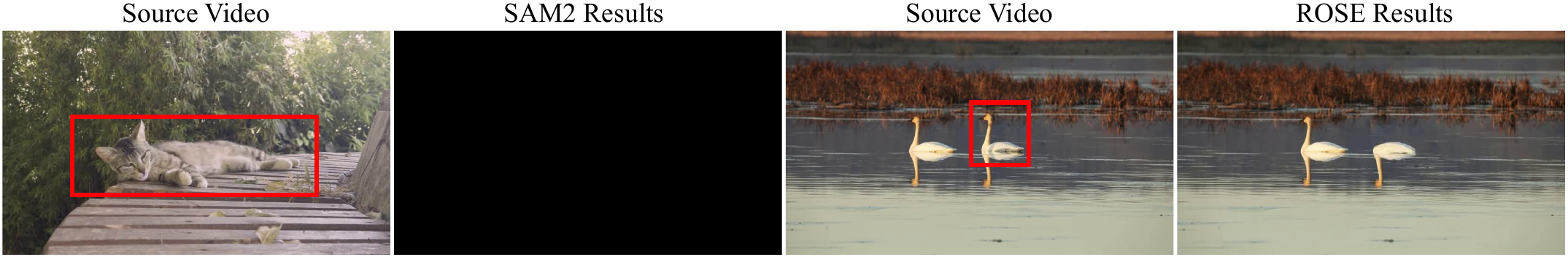}
    \vspace{-4ex}
    \caption{Failure cases of SAM-based masking and ROSE object removal, where object masks are incomplete or visual effects are not fully removed. 
    }
    \label{fig:dataset_sam_rose}
\end{figure*}

\paragraph{SAM-mask and object removal filtering.}
To obtain clean background videos, we remove objects from composite videos using ROSE~\citep{miao2025rose}. Since ROSE requires object masks as input, we generate object mask videos using SAM2~\citep{ravi2024sam} and Grounded-DINO~\citep{ren2024grounding}.  However, as shown in \cref{fig:dataset_sam_rose}, we observe that object masks are occasionally incomplete or inaccurate, leading to failure cases where ROSE does not fully remove the object or its associated visual effects. To address this issue, we apply a human filtering step to discard such low-quality samples.

\begin{figure*}[t]
    \centering
    \includegraphics[width=\textwidth]{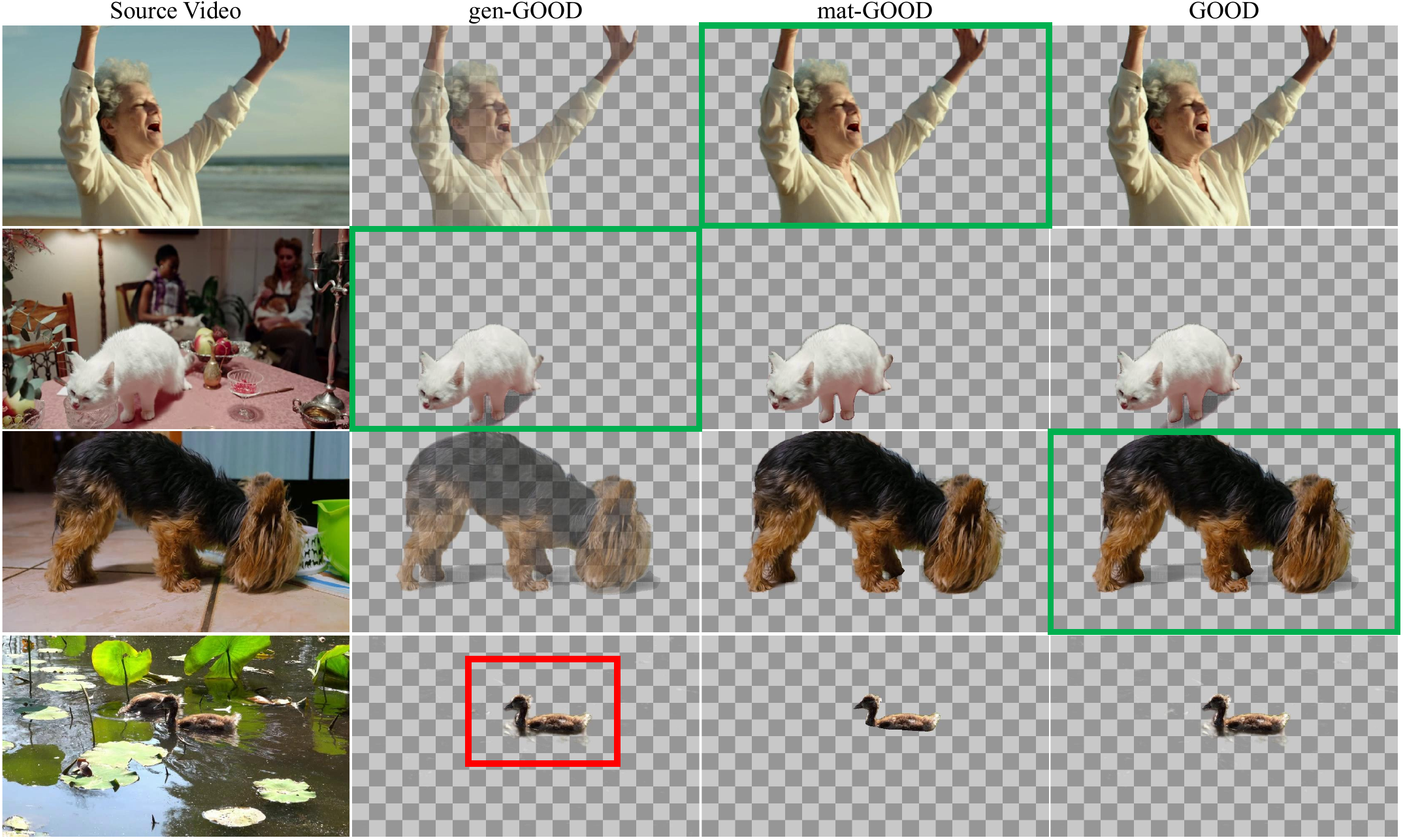}
    \vspace{-4ex}
    \caption{Examples of human annotations for foreground layer quality. The \green{green box} indicates the selected foreground layer label. The bottom row is labeled as \textit{BAD}, as it fails to capture the object's associated visual effects.}
    \label{fig:dataset_layer_label}
\end{figure*}

\paragraph{Foreground layer filtering.}
To obtain high-quality foreground layers, we combine Gen-Omnimatte~\citep{lee2025generative}, MatAnyone~\citep{yang2025matanyone}, and SAM2~\citep{ravi2024sam}, as illustrated in \cref{fig:dataset_layer_label}. While Gen-Omnimatte effectively captures visual effects, it often produces semi-transparent results for object regions. To address this issue, we incorporate SAM2 and MatAnyone to refine object opacity and recover fine-grained structures, such as hair or fur, resulting in more accurate and visually coherent foreground layers. To further ensure quality, we perform human verification on the generated results, as shown in \cref{fig:dataset_layer_label}. Each sample is manually annotated with one of four labels: \emph{GOOD}, \emph{gen-GOOD}, \emph{mat-GOOD}, and \emph{BAD}. Specifically, \emph{gen-GOOD} indicates that the Gen-Omnimatte result alone is of acceptable quality, \emph{mat-GOOD} corresponds to cases where SAM2 and MatAnyone provide better results, \emph{GOOD} denotes samples where all methods are successfully combined, and \emph{BAD} refers to low-quality outputs that are discarded.

Based on these annotations, we define the refined foreground layer and alpha matte as follows:
\begin{equation}
\begin{split}
M_{mat\_GOOD} &= max(M_{sam}, M_{mat}), \\
V_{mat} &= M_{mat\_GOOD} \times V_{rgb}, \\
V_{fg}[i,j] &=
\begin{cases}
V_{mat}[i,j], & \text{if } M_{mat\_GOOD}[i,j] > M_{gen\_GOOD}[i,j] \\
V_{fg}^{omni}[i,j], & \text{otherwise}
\end{cases} \\
M_{GOOD} &= max(M_{mat\_GOOD}, M_{gen\_GOOD}), \\
\end{split}
\label{eq:dataset_eq}
\end{equation}
where $M_{sam}$, $M_{mat}$, and $M_{gen\_GOOD}$ denote the alpha masks obtained from SAM2, MatAnyone, and Gen-Omnimatte, respectively. $V_{rgb}$, $V_{fg}^{omni}$, and $V_{fg}$ denote the composite video, the foreground layer obtained from Gen-Omnimatte, and the refined foreground layer, respectively. $M_{mat\_GOOD}$ and $M_{GOOD}$ denote the refined alpha mattes corresponding to \emph{mat-GOOD} and \emph{GOOD}, respectively.

\paragraph{Background refinement.}
For background refinement, we apply our proposed \MethodNameRefinement{} using a pretrained video inpainting model~\citep{jiang2025vace}.
As shown in \cref{tab:prompt_inp_maker}, we take the object-removed background videos previously obtained using ROSE and create a corresponding text prompt for each video.
These background videos and text prompts are then used as the source videos and textual conditioning for \MethodNameRefinement{}, respectively.
After applying \MethodNameRefinement{}, we perform human filtering to compare the object-removed backgrounds obtained from ROSE with the refined results and select the higher-quality result for each video.
Specifically, among the 3,908 background videos, 3,542 are replaced with the \MethodNameRefinement{} results, while 366 retain the original object-removed results.
The detailed algorithm and implementation are described in Sec.~\ref{sec:AGBI}.

\begin{figure*}[t]
    \centering
    \includegraphics[width=0.85\textwidth]{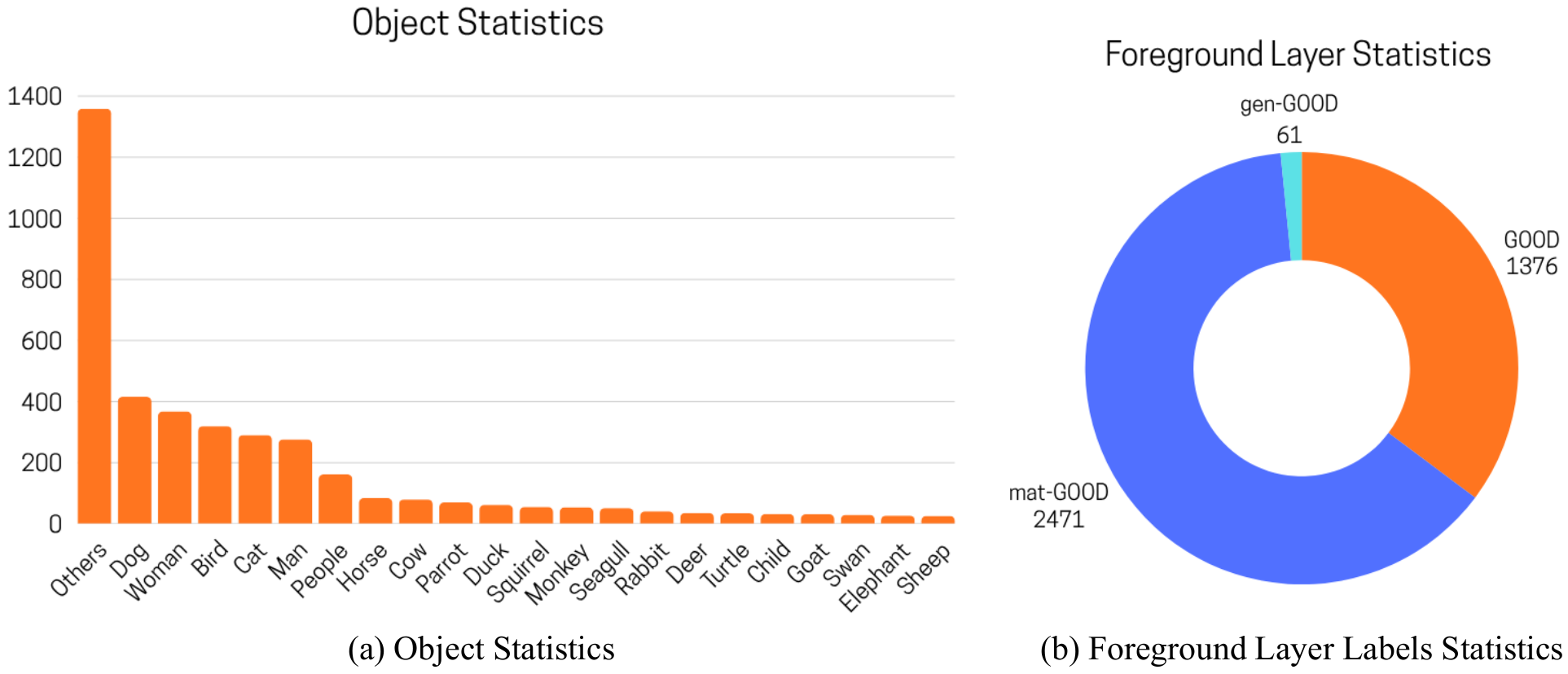}
    \vspace{-2.5ex}
    \caption{Statistics of the \DatasetName{} dataset. (a) Distribution of object categories. (b) Distribution of foreground layer labels.}
    \label{fig:dataset_stat}
\end{figure*}

\paragraph{Statistics.}
\DatasetName{} consists of 3,908 samples, covering 229 object categories. As shown in \cref{fig:dataset_stat}(a), the dataset exhibits a diverse distribution of object categories, where categories not explicitly listed are grouped into \emph{Others}. In \cref{fig:dataset_stat}(b), we present the distribution of foreground layer labels assigned via human verification, which ensures the quality of layer merging annotations.

\begin{figure*}[t]
    \centering
    \includegraphics[width=\textwidth]{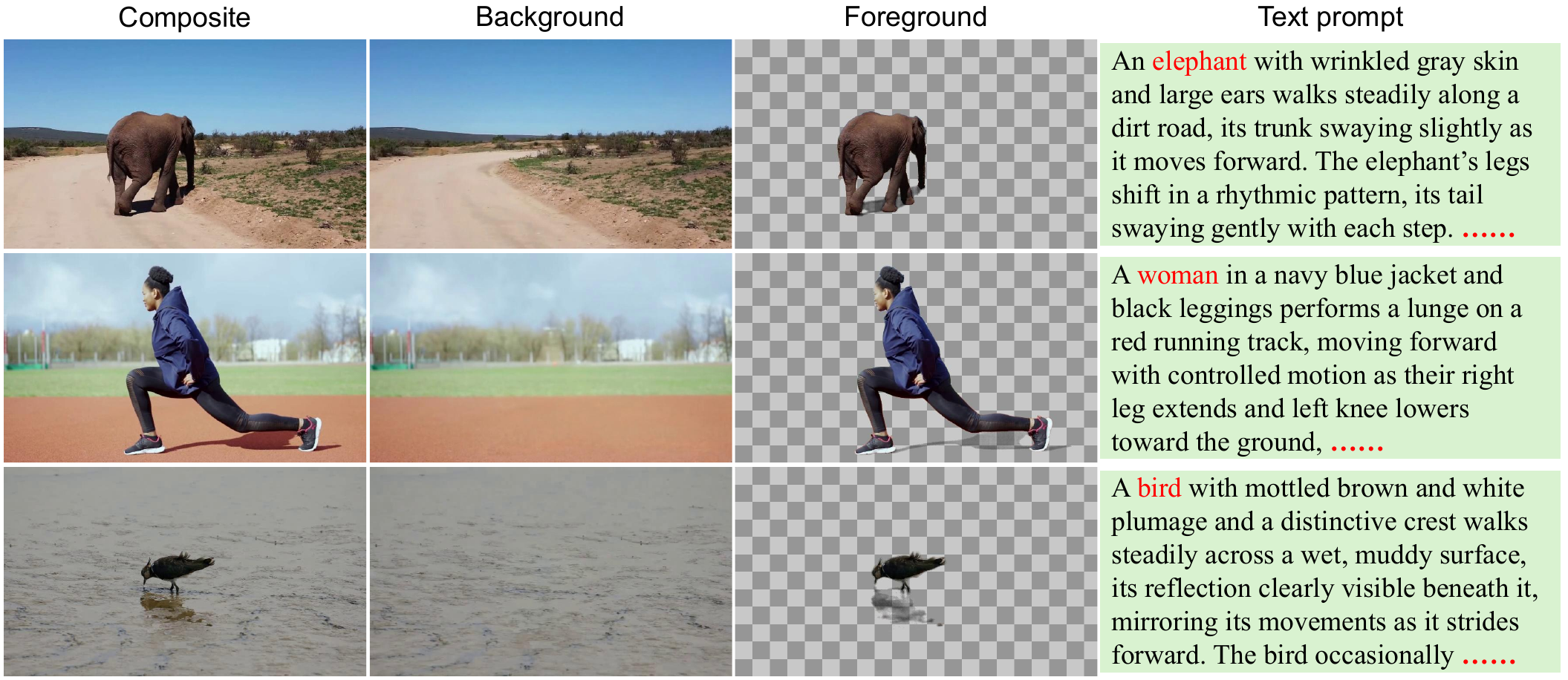}
    \vspace{-4ex}
    \caption{Additional examples from the \DatasetName{} dataset.
    }
    \label{fig:dataset_more}
\end{figure*}

As shown in \cref{fig:dataset_more}, additional examples from the dataset are presented.
More dataset examples are provided in the supplementary video.


\clearpage
\begin{algorithm}[t]
\DontPrintSemicolon
\caption{Background Refinement - Appearance-Guided Background Inpainting (AGBI)}
\label{alg:agbi}

\KwIn{
    \\
    \hspace{1em}{\scriptsize$\bullet$} Pretrained inpainting model: $v_{\theta}$\
    
    \hspace{1em}{\scriptsize$\bullet$} Target timesteps: $\{t_i\}_{i=1}^{T}
        :
        t_1 > t_2 > \cdots > t_T$;
        where $t_i\in[0,1]$\
    
    \hspace{1em}{\scriptsize$\bullet$} Source background video: $X^{\mathrm{src}}$\
    
    \hspace{1em}{\scriptsize$\bullet$} Noise strength: $\sigma_s \in[0,1]$\

    \hspace{1em}{\scriptsize$\bullet$} Text embedding: $E_t$\

    \hspace{1em}{\scriptsize$\bullet$} Object-level mask: $M$\

    \hspace{1em}{\scriptsize$\bullet$} Bounding-box mask: $B$\
    
    \hspace{1em}{\scriptsize$\bullet$} Number of initial injection steps: $N$\
}

\KwOut{Clean background video $X^{\mathrm{tar}}$}

\;
\tcp{Define start timestep using noise strength}
$T_s = T-\lfloor T\sigma_s \rfloor+1$

\;
\green{\tcp{Appearance-Guided Background Inpainting}}
$Z^{\mathrm{src}}_{0} = Encode(X^{\mathrm{src}})$

\For{$i=T_s$ \KwTo $T$}{

    \If{$i = T_s$}{
        Sample \blue{$\epsilon \sim \mathcal{N}(0,I)$}
        
        \blue{$Z^{\mathrm{tar}}_{t_{T_s}}
        =$
        $(1-t_{T_s})Z^{\mathrm{src}}_{0}
        +t_{T_s}\epsilon$ \tcc*{Define noised target latent}}
    }

    \;
    \tcp{Predict velocity}
    \eIf{\blue{$(i - T_s) < N$}}{
        $V_i = v_{\theta}(Z^{\mathrm{tar}}_{t_i},t_i,$ \blue{$B$}$, E_t)$\;
    }{
        $V_i = v_{\theta}(Z^{\mathrm{tar}}_{t_i},t_i,$ \blue{$M$}$, E_t)$\;
    }
    
    \;
    $Z^{\mathrm{tar}}_{t_{i+1}}
    =
    \mathrm{Denoise}
    \left(
        Z^{\mathrm{tar}}_{t_i},
        t_i,
        V_i
    \right)$\ \blue{\tcc*{Denoise noised target latent}}
}

$X^{\mathrm{tar}}
\gets
\mathrm{Decode}\left(Z^{\mathrm{tar}}_{t_{T+1}}\right)$\;

\Return{$X^{\mathrm{tar}}$}\;

\end{algorithm}
\begin{table}[t]
    \centering
    \caption{Hyperparameter settings for AGBI. 
    The hyperparameters can be adjusted for each video case according to the type and severity of the artifacts.}
    \label{tab:agbi_hyperparameters}
    \begin{tabular}{lc}
        \toprule
        \textbf{Hyperparameter} & \textbf{Value} \\
        \midrule
        Scheduler & UniPCMultistepScheduler \\
        Target sampling steps ($T$) & 50 \\
        Classifier-free guidance scale & 5.0 \\
        Noise strength ($\sigma_s$) & 0.8 \\
        Start timestep index ($T_s$) & 11 \\
        Flow shift & 3.0 \\
        Number of initial injection steps $N$ & 10 \\
        \bottomrule
    \end{tabular}
\end{table}
\begin{figure*}[t]
    \centering
    \includegraphics[width=\textwidth]{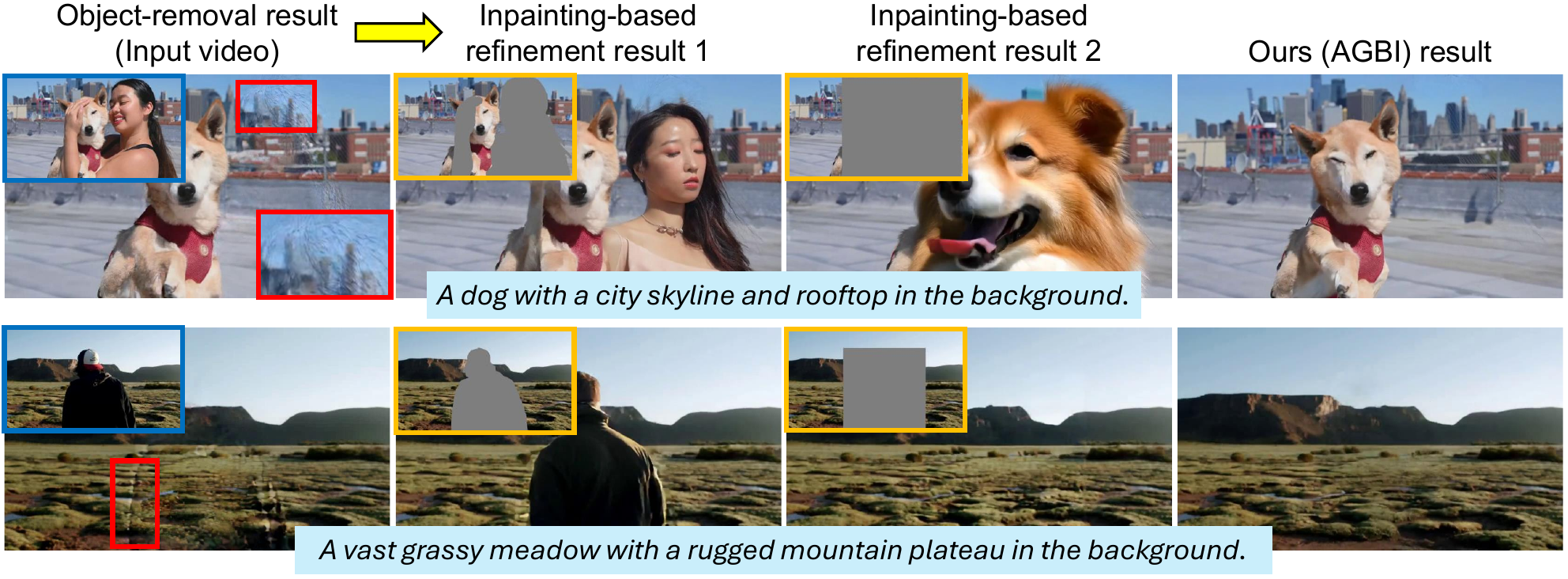}
    \vspace{-5ex}
    \caption{
    Comparison of background refinement with inpainting-based results.
    The input videos are background reconstruction results obtained using an object removal model. 
    The inpainting-based results use object-level and bounding-box masking, respectively. The last column shows our \MethodNameRefinement{} results. 
    The blue and orange boxes denote the composite and masked video, respectively.
    }
    \label{fig:AGBI_introduce}
\end{figure*}

\section{Details of Background Refinement}
\label{sec:AGBI}
Alg.~\ref{alg:agbi} illustrates the overall procedure of \MethodNameRefinement{}.
Existing object removal methods can effectively
remove the target object and its associated visual effects, but may leave
residual artifacts in the reconstructed background. 
A straightforward solution is to refine these regions with a pretrained inpainting model using the foreground alpha matte as a binary mask, but naive inpainting can lead to undesirable results. 
As shown in \cref{fig:AGBI_introduce} and~\ref{fig:agbi_more_qual}, object-level masking can preserve the surrounding background appearance but often generates unintended structures following the shape of the masked region,
whereas bounding-box masking can more effectively mitigate residual
artifacts but often alters the appearance of the background within the
masked region.

To address this issue, \MethodNameRefinement{} combines the
inpainting prior with source background appearance guidance through a two-stage
masking strategy.
Given an object-removed background video obtained
from the object removal model, we first construct the object-level mask $M$ from
the alpha matte corresponding to the removed object and its associated
visual effect in the \DatasetName{} dataset. We then obtain the bounding-box
mask $B$ by taking the bounding box of the object-level mask,
providing a broader region for the initial refinement.

Given the object-removed background video $X^{\mathrm{src}}$, we encode
it into the source latent $Z_0^{\mathrm{src}}$. Rather than initializing
the refinement process from pure noise, we construct a partially noised
target latent according to a prescribed noise strength $\sigma_s \in [0,1]$:
\begin{equation}
Z_{t_{T_s}}^{\mathrm{tar}}
=
(1-t_{T_s})Z_0^{\mathrm{src}} + t_{T_s}\epsilon,
\qquad
\epsilon\sim\mathcal{N}(0,I),
\label{eq:agbi_source_noise}
\end{equation}
where $T_s = T-\lfloor T\sigma_s \rfloor+1$ and 
$\{t_i\}^{T}_{i=T_{s}}: t_{T_s} > t_{T_s+1} > \cdots > t_T; \text{ } t_i \in [0,1]$.
This initialization retains useful appearance and
contextual information from the object-removed background while
providing sufficient noise for subsequent refinement.

Starting from $Z_{t_{T_s}}^{\mathrm{tar}}$, we perform denoising using the
pretrained video inpainting model. We divide the denoising process into
two stages. During the first $N$ denoising steps, we apply the
bounding-box mask $B$ to provide a sufficiently broad region
for the inpainting prior to suppress residual artifacts and reconstruct
a coherent background. For the remaining denoising steps, we switch to
the object-level mask $M$. This restricts further modification
to the region associated with the removed object and its visual effect,
allowing the inpainting prior to refine the remaining artifacts while
preserving the appearance and fine-grained details of the surrounding
background.

Formally, let $i\in[T_s,T]$ denote the current denoising step. The mask used at each step
is defined as
\begin{equation}
\mathcal{M}_i =
\begin{cases}
B, & (i-T_s) < N,\\
M, & (i-T_s) \geq N.
\end{cases}
\end{equation}
The corresponding velocity is obtained from the pretrained inpainting
model as
\begin{equation}
V_i =
v_{\theta}
\left(
Z_{t_i}^{\mathrm{tar}},
t_i,
\mathcal{M}_i,
E_t
\right),
\end{equation}
where $v_{\theta}$ and $E_t$ denote the velocity field of the pretrained
inpainting model and text embedding, respectively. 
The target latent is then progressively denoised
toward $t=0$, producing the refined background video
$X^{\mathrm{tar}}$.

The two-stage masking strategy provides complementary benefits:
the initial bounding-box refinement broadly suppresses residual
artifacts, while the subsequent object-level refinement minimizes
unnecessary changes to the surrounding background. Together with the
partially noised latent initialization, \MethodNameRefinement{} produces
a refined background that effectively removes residual artifacts while
retaining the appearance of the original background. 
Moreover, \MethodNameRefinement{} is not inherently tied to our pipeline and can potentially be applied to background videos produced by other video object removal or video layer decomposition methods, indicating its broader applicability for background refinement.
More discussion and ablation studies are provided in Sec.~\ref{sec:AGBI_discussion} and~\ref{sec:AGBI_ablation}, respectively.

\paragraph{Hyperparameters.}
\cref{tab:agbi_hyperparameters} summarizes the detailed hyperparameters
used for \MethodNameRefinement{}. For background refinement, we employ VACE-1.3B~\citep{jiang2025vace} as the pretrained inpainting model. In addition, we use the following negative prompt for CFG: \texttt{"Bright tones, overexposed, static, blurred details, subtitles, style, works, paintings, images, static, overall gray, worst quality, low quality, JPEG compression residue, ugly, incomplete, extra fingers, poorly drawn hands, poorly drawn faces, deformed, disfigured, misshapen limbs, fused fingers, still picture, messy background, three legs, many people in the background, walking backwards, changing color"}. The background refinement process takes about 3 minutes per video on a single NVIDIA A100 GPU.

\section{Details of LayeredVid-Benchmark}
\label{sec:benchmark}
\begin{figure*}[t]
    \centering
    \includegraphics[width=\textwidth]{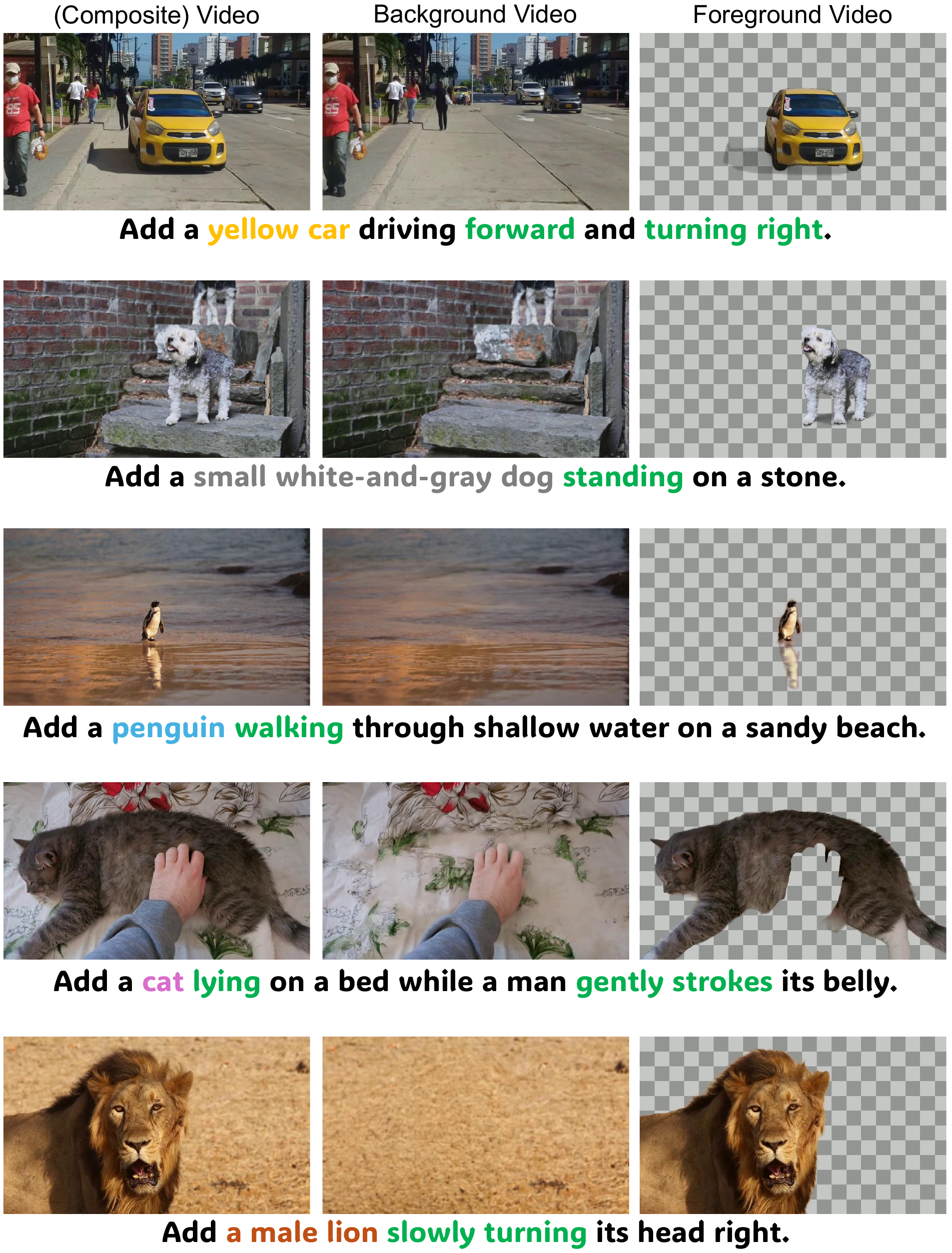}
    \vspace{-4ex}
    \caption{Some examples from the LayeredVid-Benchmark.
    }
    \label{fig:benchmark_examples}
\end{figure*}

\cref{fig:benchmark_examples} illustrates some examples of LayeredVid-Benchmark. There is a lack of publicly available benchmarks for video object insertion. Moreover, existing benchmarks for video layer decomposition, such as Movies~\citep{lin2023omnimatterf} and Kubric~\citep{wu2022d}, primarily evaluate background reconstruction and do not provide ground-truth foreground layers. Consequently, prior layer decomposition methods have largely been evaluated only in terms of background reconstruction quality.

Although video object insertion and video layer decomposition represent different tasks, they provide complementary perspectives on layered video generation. 
Video object insertion focuses on generating a visually plausible foreground object and naturally compositing it into a target scene. 
In contrast, video layer decomposition aims to recover foreground and background layers from a composite video, allowing direct evaluation against ground-truth layers when available. 

We therefore design \emph{LayeredVid-Benchmark} to cover both aspects, enabling the evaluation of perceptual quality in both video object insertion and video layer decomposition.
We construct the benchmark by randomly sampling 65 videos from \DatasetName{} as a held-out test set from the training split, considering diverse objects and scenarios to cover a wide range of realistic scenarios. Each video is 5 seconds long and contains 81 frames, comprising 35 videos with object-only foreground content and 30 videos containing both object appearance and associated visual effects such as shadows and reflections.
As shown in \cref{fig:benchmark_examples}, the benchmark includes diverse examples with varying object appearances, motion, background scenes, and scene--object interactions.
This benchmark enables evaluation not only of whether an inserted object is visually consistent with the target scene, but also of whether the corresponding foreground and background layers plausibly capture the object appearance, visual effects, and underlying scene content, respectively.

We believe that LayeredVid-Benchmark provides a more comprehensive evaluation protocol for layered video generation and decomposition, facilitating future research on explicit layer modeling and practical video editing.
Further discussion is provided in the Sec.~\ref{sec:benchmark_discussion}.

\section{Details of Disentangled Timestep Sampling}
Within the rectified flow framework, we use disentangled timestep sampling to stabilize joint optimization of the RGB and RGBA branches.
Each branch evolves at its own noise level, which balances the
learning dynamics of RGB appearance and RGBA layer reconstruction while
preserving effective cross-branch interaction. Concretely, we sample
independent timesteps $t_x, t_y \sim \mathcal{U}(0,1)$ for the RGB and
RGBA branches and train the model to predict the corresponding
velocities. The overall objective is the sum of the RGB and RGBA
velocity-prediction losses, with equal weighting in practice.

\begin{figure*}[t]
    \centering
    \includegraphics[width=\textwidth]{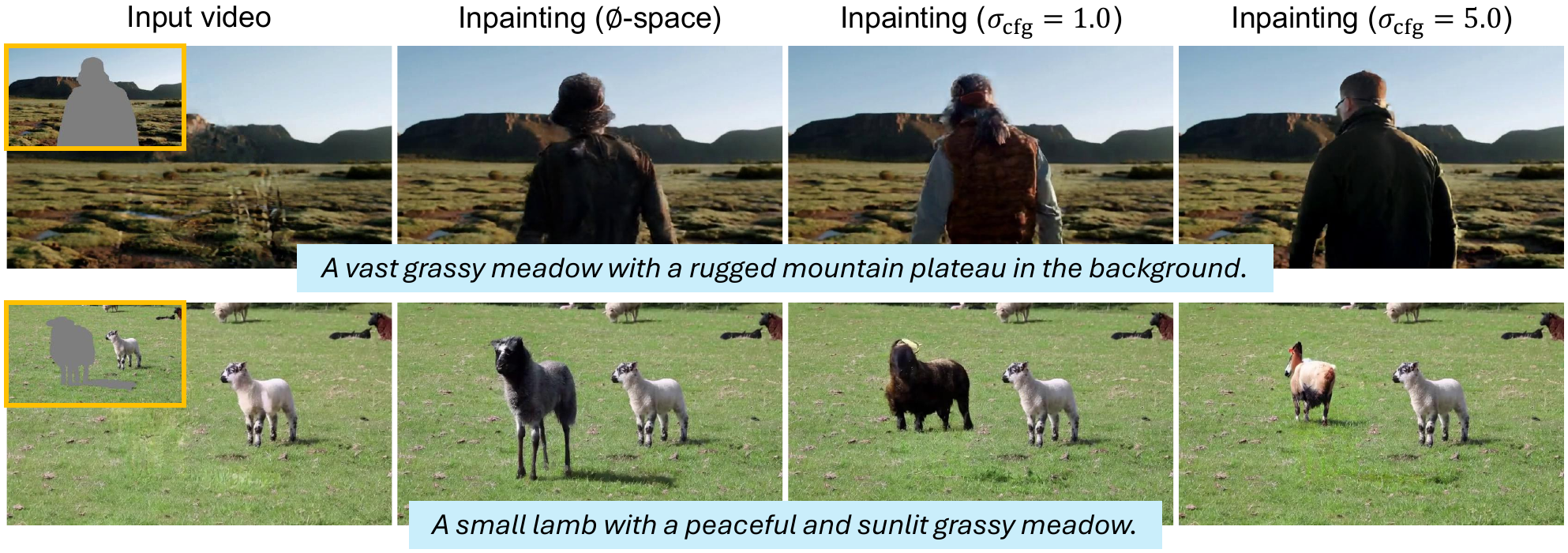}
    \vspace{-5ex}
    \caption{
    Qualitative analysis of the effects of the object-level mask and text prompt in video inpainting.
    The input video is the source background video obtained from the object removal model.
    The orange box denotes the masked video condition for inpainting.
    }
    \label{fig:AGBI_null_space}
\end{figure*}

\section{Discussion: Inpainting-based Background Refinement}
\label{sec:AGBI_discussion}
Existing video object removal methods are designed to remove target objects and their associated visual effects, but can still leave residual artifacts, such as afterimages and ghosting, in the reconstructed background. 
In contrast, video inpainting models take a masked video as input and generate plausible content to fill the masked regions. 
Therefore, a straightforward way to refine the artifacts remaining after object removal is to use the object-level alpha matte of the removed object and visual effect as the binary mask for inpainting.

However, as shown in \cref{fig:AGBI_introduce} and~\ref{fig:agbi_more_qual}, directly applying object-level masking to a video inpainting model can produce undesirable results, where the generated content follows the shape of the mask more strongly than the provided text prompt.
To investigate this behavior, we analyze the relative influence of the mask and text conditioning under three settings: $\emptyset$-space (text unconditioning), CFG scale $\sigma_{\mathrm{cfg}}=1.0$ and $5.0$. We use VACE-1.3B~\citep{jiang2025vace} as the inpainting model and the object-removed background videos produced by ROSE~\citep{miao2025rose} as inputs.

As shown in \cref{fig:AGBI_null_space}, with $\emptyset$-space, the model often generates object-like structures that closely follow the geometry of the object-level mask. 
This indicates that the inpainting-prior learns a strong spatial preference for completing masked regions according to the shape of the mask. 

Interestingly, utilizing text conditioning with either a $\sigma_{\mathrm{cfg}}=1.0 \text{ or }5.0$ does not substantially alter this behavior: the generated content remains strongly correlated with the shape of the masked region rather than the text prompt. 
This suggests that, although the text prompt provides global semantic guidance, the mask structure serves as a stronger spatial cue when reconstructing the masked region. 

Therefore, a broader mask is preferable to a strictly object-level mask, as it reconstructs plausible background content and reduces the tendency to generate object-like structures that follow the mask shape.
Consequently, bounding-box masking is more effective at both suppressing residual artifacts and reconstructing plausible background content. However, its wider masked area also excludes the original background appearance from the inpainting input, resulting in unnecessary changes to existing background details and textures.

These observations motivate our \MethodNameRefinement{} design. Rather than relying solely on the inpainting prior, \MethodNameRefinement{} provides explicit appearance guidance from the source background video and employs a two-stage masking strategy. The initial bounding-box mask enables broad artifact refinement, while the subsequent object-level mask restricts further modification to the removed region and preserves the surrounding background appearance. This combination allows \MethodNameRefinement{} to exploit the artifact-removal capability of the inpainting prior while reducing its sensitivity to mask shape and retaining useful appearance information from the original background.

\section{Discussion: LayeredVid-Benchmark}
\label{sec:benchmark_discussion}
Our LayeredVid-Benchmark addresses an important gap in the evaluation of layered video generation.
For video object insertion, the lack of publicly available benchmarks makes it difficult to systematically compare existing methods.
Meanwhile, existing benchmarks for video layer decomposition, such as Movies~\citep{lin2023omnimatterf} and Kubric~\citep{wu2022d}, primarily focus on background reconstruction, as ground-truth foreground layers are not provided.
As a result, existing evaluation protocols cannot directly assess the quality of the foreground layer and its associated visual effects.

LayeredVid-Benchmark can evaluate both video object insertion and video layer decomposition by providing the composite video and its corresponding foreground and background layers. 
It assesses the perceptual quality of the inserted video as well as the individual layers, including object appearance, visual effects, and underlying scene content. 
This allows us to examine whether a method produces not only a plausible composite video but also visually consistent and meaningful layers.

However, we acknowledge that the background and foreground layers in our benchmark are not manually annotated by humans and instead rely on pretrained models within our data collection pipeline.
Therefore, the benchmark may not be regarded as providing perfectly accurate ground truth.
Nevertheless, we believe that this approach offers a practical trade-off between annotation cost and evaluation coverage, enabling perceptual evaluation of both insertion and decomposition across diverse real-world scenarios.
Future research could further improve this benchmark by developing more reliable layer annotations and validation procedures that incorporate annotations generated by humans.

\begin{figure*}[t]
    \centering
    \includegraphics[width=\textwidth]{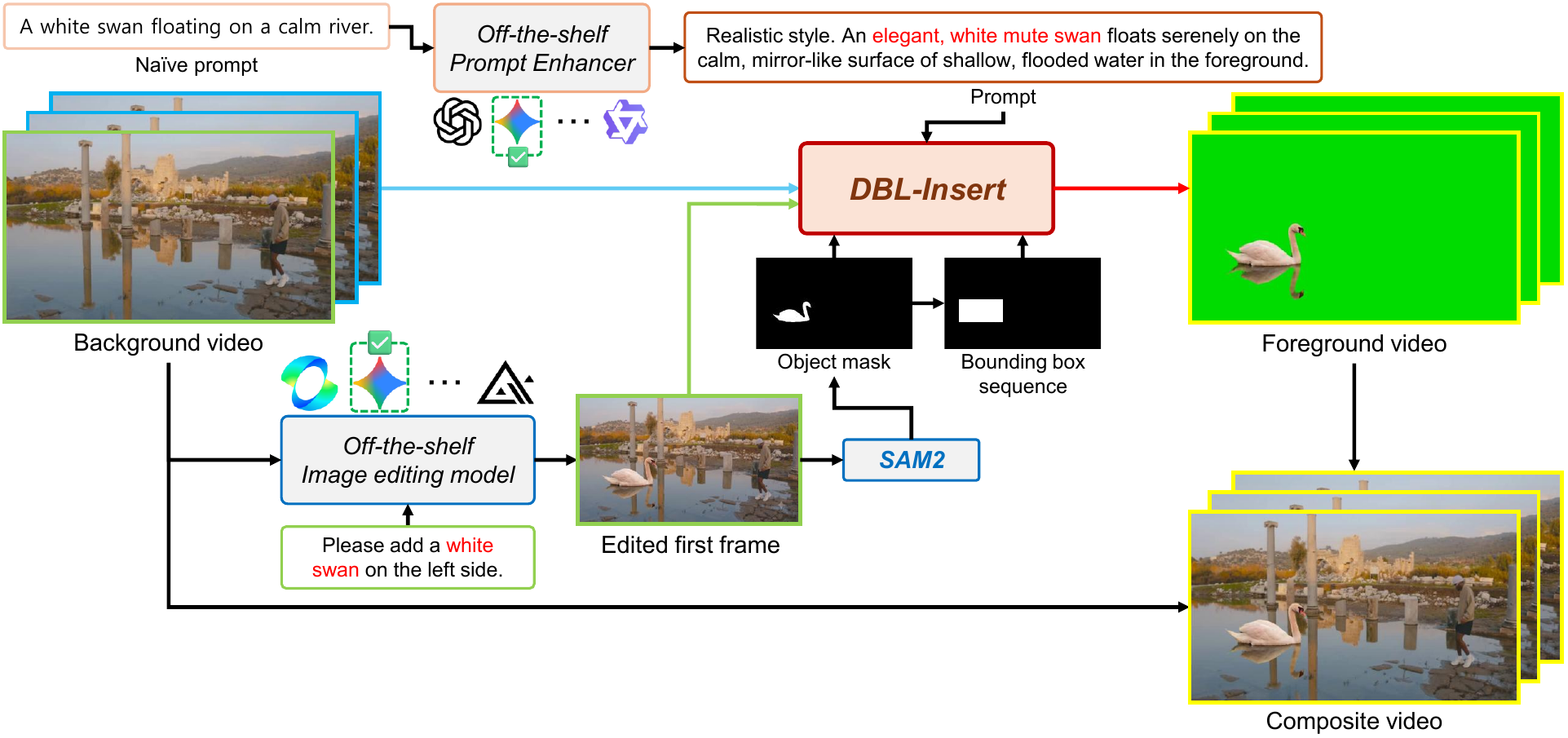}
    \vspace{-5ex}
    \caption{Inference process of \MethodNameInsert{} for creating custom video layered object insertion examples.}
    \label{fig:inference_process_insert}
\end{figure*}
\begin{figure*}[t]
    \centering
    \includegraphics[width=\textwidth]{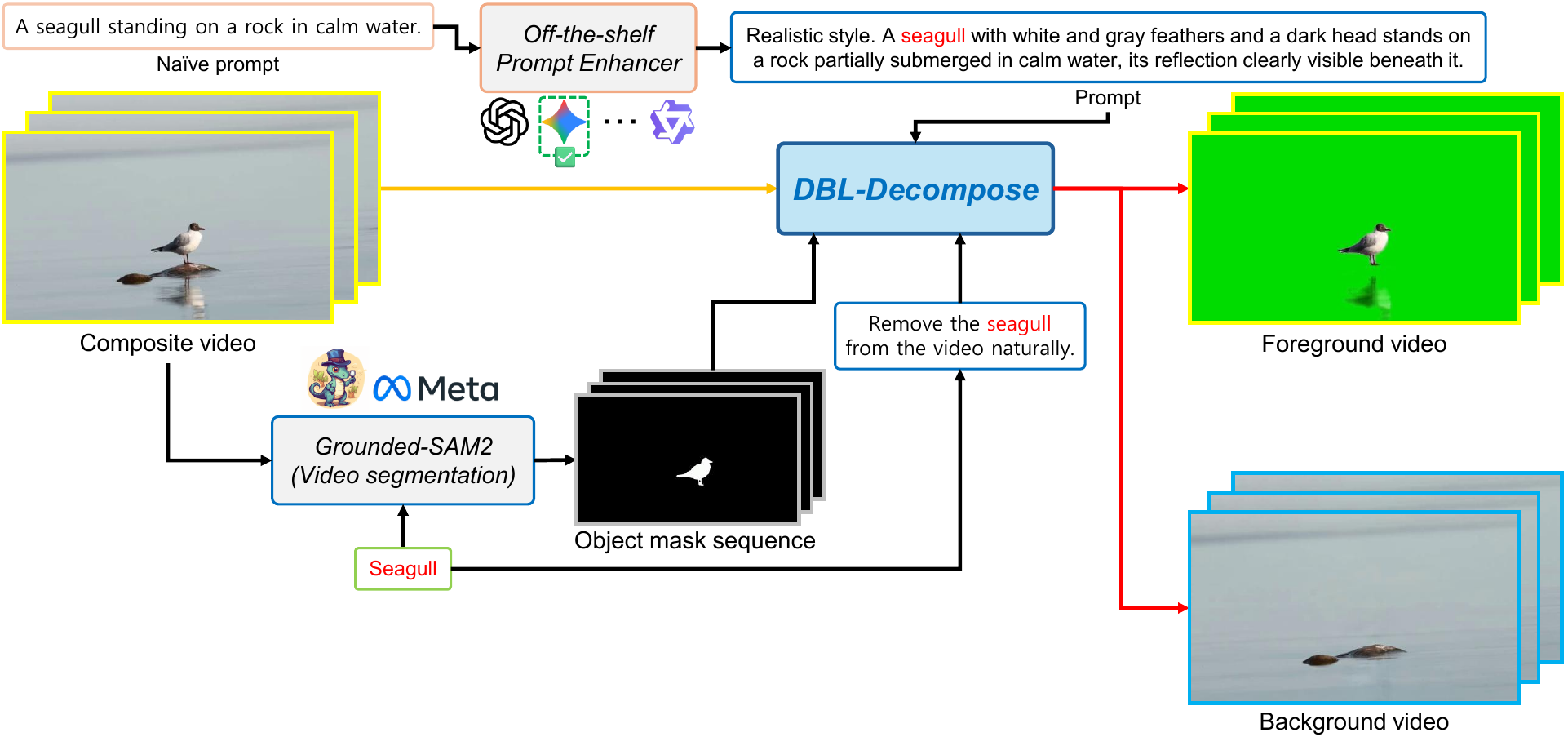}
    \vspace{-5ex}
    \caption{Inference process of \MethodNameDecompose{} for creating custom video layer decomposition examples.}
    \label{fig:inference_process_decompose}
\end{figure*}

\section{Detailed Inference process of \MethodNameInsert{} on a custom example}
\cref{fig:inference_process_insert} and~\ref{fig:inference_process_decompose} illustrate the inference procedure of \MethodNameInsert{} and \MethodNameDecompose{} for creating a custom video layered object insertion and layer decomposition example, respectively.

First, we use an off-the-shelf image editing model to insert the desired object into the first frame of the given background video. In our implementation, we primarily use Gemini~\citep{Nano-Banana-2} or Kling~\citep{kling}, which allows the object to be specified using either a reference image or a text prompt.
Second, we extract the mask of the inserted object using SAM2. We then construct a sequence of bounding boxes that provides coarse spatial guidance for the object's motion throughout the video.
Third, we prepare an initial text prompt describing the desired motion of the inserted object in the given background video. We then use an off-the-shelf prompt enhancement model to obtain a semantically enriched prompt. In our implementation, we primarily use Gemini and ChatGPT~\citep{gpt-5.6} for prompt enhancement.
Finally, we provide all the prepared conditions to \MethodNameInsert{} to generate the foreground layer. We then alpha-blend the generated foreground layer with the given background video to produce the final composite video.

For \MethodNameDecompose{}, we utilize Grounded-SAM2 to obtain the object mask sequence, while keeping the same text prompt enhancement process. We then provide all the prepared conditions to \MethodNameDecompose{} to generate the foreground and background layers.

\section{Social Impacts}
Our work has the potential to facilitate practical video editing in applications such as filmmaking, advertising, and visual effects production. By explicitly modeling foreground layers, our framework enables flexible manipulation of objects and visual effects, while supporting both layered video object insertion and layer decomposition. This can reduce the cost and effort required for conventional VFX workflows and make advanced video editing more accessible.

At the same time, the ability to realistically insert or remove visual content may be misused to create misleading or deceptive media. We therefore encourage the use of our models for authorized and clearly disclosed editing purposes. For real-world deployment, content provenance, watermarking, and appropriate disclosure mechanisms can further help distinguish generated or edited content from authentic footage.


\clearpage

\begin{table*}[t]
    \centering
    \caption{Quantitative Comparison: Quality of inserted foreground layers. `VACE-Inp' denotes an inpainting model using pre-trained VACE, and `VACE-F' denotes VACE fine-tuned on the \DatasetName{} dataset for the video layered object insertion task.
    }
    \resizebox{\textwidth}{!}{
        \begin{tabular}{l|c|cccc|cc}
        \toprule
        Model & 
        ViCLIP-T $\uparrow$ & 
        \makecell{Subject \\ Consistency $\uparrow$} & 
        \makecell{Motion \\ Smoothness $\uparrow$} &
        Aesthetic $\uparrow$ & 
        \makecell{Imaging \\ Quality $\uparrow$} &
        CLIP-I $\uparrow$ & 
        DINO-I $\uparrow$ \\
        \midrule
        AnyV2V + Gen-Omnimatte
        & 12.273 & 0.902 & 0.996 & 0.345 & 0.458 & 0.641 & 0.274 \\
        
        ReVideo + Gen-Omnimatte
        & 14.451 & 0.911 & \textbf{0.997} & 0.363 & 0.515 & 0.663 & 0.395 \\

        VACE-Inp + Gen-Omnimatte
        & 13.880 & 0.921 & \textbf{0.997} & 0.358 & 0.507 & 0.667 & 0.408 \\

        VACE-F
        & 17.496 & 0.879 & 0.984 & 0.361 & 0.594 & 0.683 & 0.468 \\
        

        \cellcolor{blue!15}\MethodNameInsert{} (RGBA branch)
        & \cellcolor{blue!15}\textbf{17.520} & \cellcolor{blue!15}\textbf{0.951} & \cellcolor{blue!15}0.992 & \cellcolor{blue!15}\textbf{0.409} & \cellcolor{blue!15}\textbf{0.633} & \cellcolor{blue!15}\textbf{0.686} & \cellcolor{blue!15}\textbf{0.472} \\
        
        \bottomrule
    \end{tabular}
    }
    \label{tab:compare_voli_models}
\end{table*}
\begin{table}[t]
    \centering
    \caption{Additional FID ($\downarrow$) comparison of background/foreground reconstruction.
    }
    \small
    \resizebox{0.8\linewidth}{!}{
    \begin{tabular}{l|c|c|c}
        \toprule
        Model & Kubric (BG) & Movie (BG) & LayeredVid-Benchmark (FG) \\
        \midrule
        Gen-Omnimatte & 16.598 & 16.344 & 60.826 \\
        OmnimatteZero & 96.077 & 36.428 & 46.882 \\
        \rowcolor{blue!15}
        \MethodNameDecompose{} & \textbf{15.815} & \textbf{14.364} & \textbf{32.967} \\
        \bottomrule
    \end{tabular}
    }
    \label{tab:fid_comparison}
\end{table}
\begin{table}[h]
    \centering
    \small
    \caption{Foreground layer reconstruction evaluated on LayeredVid-Benchmark.}
    \resizebox{0.6\linewidth}{!}{
        \begin{tabular}{l|ccc}
        \toprule
        Method & PSNR$\uparrow$ & LPIPS$\downarrow$ & SSIM$\uparrow$ \\
        \midrule
        Generative Omnimatte & 27.720 & 0.235 & 0.901 \\
        OmnimatteZero & 23.550 & 0.099 & 0.917 \\
        \cellcolor{blue!15}\MethodNameDecompose{} & \cellcolor{blue!15}\textbf{29.160} & \cellcolor{blue!15}\textbf{0.053} & \cellcolor{blue!15}\textbf{0.938} \\
        \bottomrule
        \end{tabular}
    }
    \label{tab:omnimatte_fg_comparison}
\end{table}

\begin{table}[t]
    \centering
    \caption{User study on video object insertion conducted with 20 participants over 10 videos. }
    \small
    \resizebox{0.8\linewidth}{!}{
    \begin{tabular}{l|cccc}
        \toprule
        Model &
        \makecell{Subject \\ Consistency$\uparrow$} &
        \makecell{Insertion \\ Rationality$\uparrow$} &
        \makecell{Prompt \\ Alignment$\uparrow$} &
        \makecell{Overall \\ Quality$\uparrow$} \\
        \midrule
        AnyV2V & 1.0\% & 1.0\% & 1.5\% & 1.5\% \\
        ReVideo & 0.5\% & 1.0\% & 1.0\% & 1.5\% \\
        VACE-Inp & 17.5\% & 10.0\% & 21.0\% & 15.5\% \\
        VACE-C & 15.0\% & 11.5\% & 18.5\% & 13.0\% \\
        VACE-F & 9.5\% & 8.5\% & 7.5\% & 7.0\% \\
        \rowcolor{blue!15}
        \MethodNameInsert{} & \textbf{56.5\%} & \textbf{68.0\%} & \textbf{50.5\%} & \textbf{61.5\%} \\
        \bottomrule
    \end{tabular}
    }
    \label{tab:user_object_insertion}
\end{table}
\begin{table}[t]
    \centering
    \caption{User study on video layer decomposition conducted with 20 participants over 10 videos.}
    \small
    \resizebox{0.55\linewidth}{!}{
    \begin{tabular}{l|cc}
        \toprule
        Model &
        \makecell{Background \\ Quality$\uparrow$} &
        \makecell{Foreground \\ Quality$\uparrow$} \\
        \midrule
        Gen-Omnimatte & 8.0\% & 19.0\% \\
        OmnimatteZero & 20.5\% & 6.0\% \\
        \rowcolor{blue!15}
        \MethodNameDecompose{} & \textbf{71.5\%} & \textbf{75.0\%} \\
        \bottomrule
    \end{tabular}
    }
    \label{tab:user_layer_decomposition}
\end{table}

\section{More Quantitative Results: Inserted Foreground Layer}
To evaluate the quality of the inserted foreground layer, we conduct quantitative experiments focusing specifically on RGBA layer outputs. Since AnyV2V, ReVideo, and VACE-Inp do not explicitly generate foreground layers, we extract RGBA layers using an off-the-shelf \emph{Omnimatte}-based model~\citep{lee2025generative}. Table~\ref{tab:compare_voli_models} presents results on the RGBA foreground layers of inserted objects.
Existing methods lack explicit supervision for layer representations, limiting their ability to capture visual effects during video object insertion. Furthermore, reliance on \emph{Omnimatte}-based extraction introduces additional limitations, such as semi-transparent object artifacts and difficulty in separating complex effects like fire and smoke. Although VACE-F benefits from dataset supervision, its single-branch design requires simultaneously modeling object appearance, scene interaction, and visual effects within a single branch, which can degrade subject consistency.

In contrast, our dual-branch design disentangles appearance and layer modeling, enabling more effective learning of both object structure and visual effects. As a result, \MethodNameInsert{} produces higher-quality and more realistic foreground layers, achieving superior performance in quantitative evaluation.

\section{More Quantitative Results: Foreground/Background Reconstruction}
Reconstruction-based \emph{omnimatte} methods are explicitly optimized for recovering the original background, which naturally favors pixel-wise metrics such as PSNR and SSIM. In contrast, our method focuses on generating perceptually plausible backgrounds and foregrounds. 
Moreover, for foreground reconstruction, our method also preserves clean and precise object boundaries while faithfully recovering object-associated effects, such as shadows, resulting in more visually coherent foreground layers (\cref{tab:fid_comparison}, \ref{tab:omnimatte_fg_comparison} and \cref{fig:DBLDecomp_Qual}).
Overall, our method achieves the best pixel-level reconstruction performance for the foreground, while offering superior perceptual fidelity for both foreground and background, as evidenced by consistently lower FID and more visually plausible results.

\section{More Quantitative Results: User Studies}

\begin{figure*}[t]
    \centering
    \includegraphics[width=\textwidth]{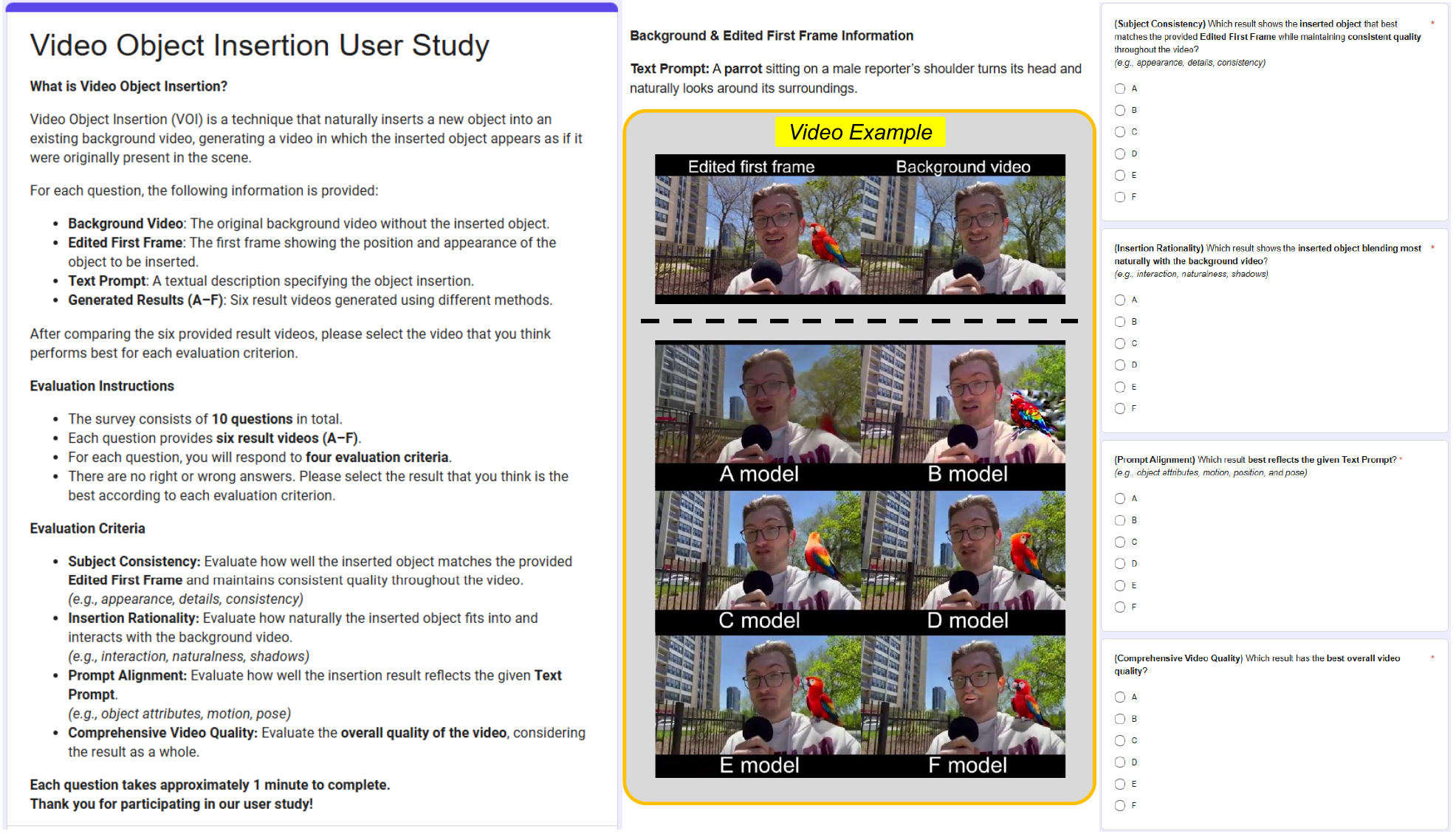}
    \vspace{-5ex}
    \caption{User study interface for video object insertion.
    }
    \label{fig:user_study_insert}
\end{figure*}

\begin{figure*}[t]
    \centering
    \includegraphics[width=\textwidth]{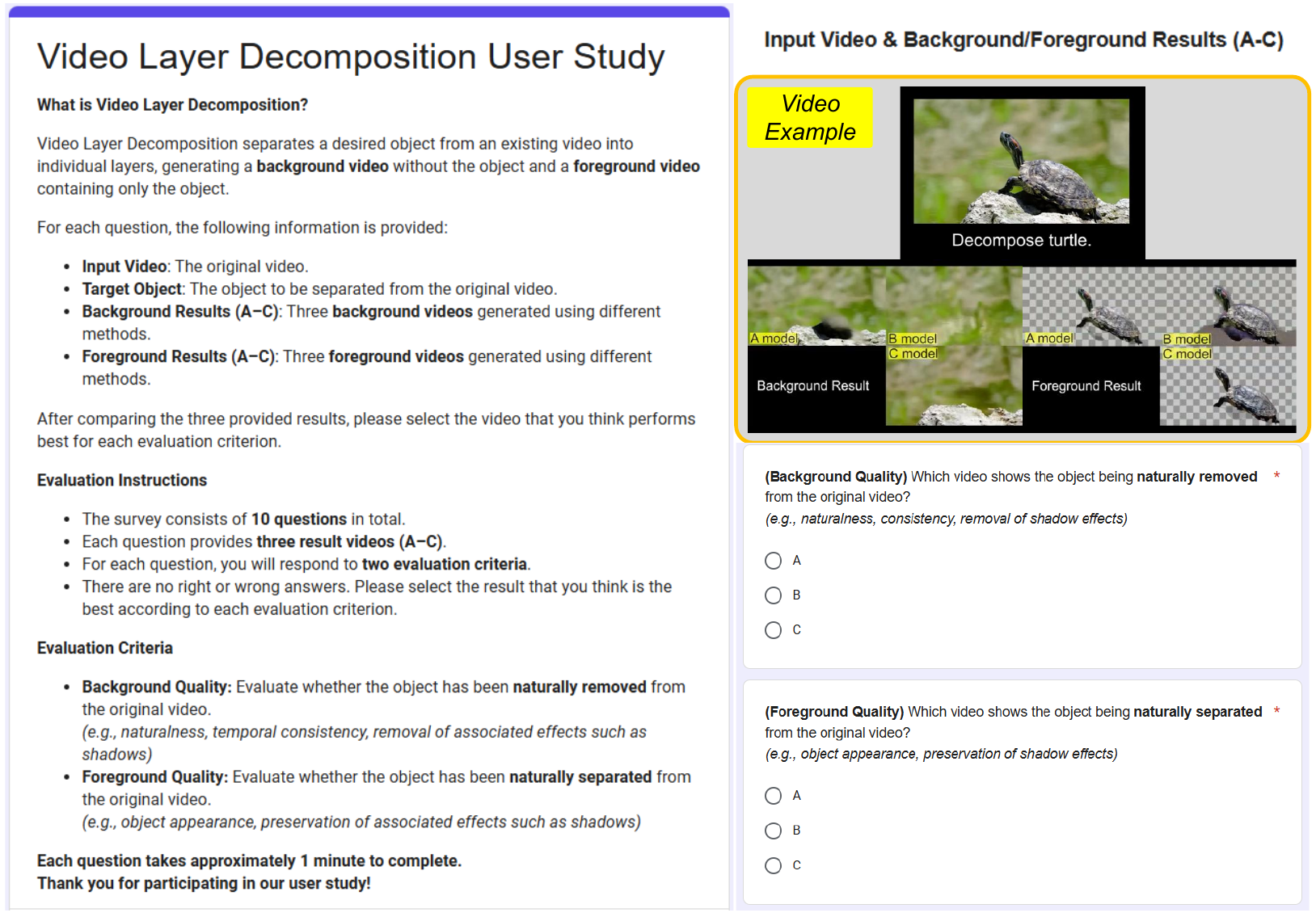}
    \vspace{-5ex}
    \caption{User study interface for video layer decomposition.
    }
    \label{fig:user_study_decompose}
\end{figure*}

To further evaluate perceptual quality, we conduct additional user studies on both video object insertion and layer decomposition.
We conduct the evaluation with 20 participants on 10 held-out in-the-wild videos from \DatasetName{}, with manually curated evaluation conditions such as edited first frames and bounding boxes.
The videos cover diverse object categories and visual effects (e.g., shadows and reflections), and scene configurations to provide a comprehensive evaluation. 
\cref{fig:user_study_insert} and~\ref{fig:user_study_decompose} illustrate the interfaces used in our user study for video object insertion and video layer decomposition, respectively.

For video object insertion, participants are asked to compare the generated videos from different methods in terms of \emph{subject consistency}, \emph{insertion rationality}, \emph{prompt alignment}, and \emph{overall quality}:
\begin{enumerate}
    \item \textbf{Subject Consistency}: Consistency of the inserted object with the provided edited first frame throughout the video, including its appearance and details.

    \item \textbf{Insertion Rationality}: Naturalness of the inserted object within the background scene, including scene interaction and shadows.

    \item \textbf{Prompt Alignment}: Degree to which the insertion result follows the given text prompt, including object attributes, motion, and pose.

    \item \textbf{Comprehensive Video Quality}: Overall visual quality of the generated video.
\end{enumerate}

As shown in \cref{tab:user_object_insertion}, \MethodNameInsert{} is consistently preferred over all competing methods across every evaluation criterion, demonstrating superior insertion quality and better alignment with the intended editing objective.

For layer decomposition, participants evaluate the perceptual quality of both reconstructed background and foreground layers:
\begin{enumerate}
    \item \textbf{Background Quality}: How naturally the object is removed from the original video (e.g., visual naturalness, temporal consistency, and removal of shadow effects).

    \item \textbf{Foreground Quality}: How naturally the object is separated from the original video (e.g., object appearance and preservation of visual effects such as shadows).
\end{enumerate}

As shown in \cref{tab:user_layer_decomposition}, \MethodNameDecompose{} receives the highest preference for both background and foreground reconstruction by a large margin, indicating that our explicit layer modeling produces more visually plausible decomposed layers than existing methods.


\clearpage

\begin{table}[t]
    \centering
    \caption{Effect of refined background datasets on background layer reconstruction performance.}
    \small
    \resizebox{0.95\linewidth}{!}{
    \begin{tabular}{l|cccc|cccc}
        \toprule
        \multirow{2}{*}{Training Dataset}
        & \multicolumn{4}{c|}{Movie} & \multicolumn{4}{c}{Kubric} \\
        &
        PSNR$\uparrow$ &
        LPIPS$\downarrow$ &
        SSIM$\uparrow$ &
        FID$\downarrow$ &
        PSNR$\uparrow$ &
        LPIPS$\downarrow$ &
        SSIM$\uparrow$ &
        FID$\downarrow$ \\
        \midrule

        w/o Refined BG Dataset &
        33.09 & 0.025 & 0.962 & 14.527 &
        38.22 & 0.021 & 0.979 & 16.340 \\

        \rowcolor{blue!15}
        w/ Refined BG Dataset &
        \textbf{33.42} & \textbf{0.017} & \textbf{0.971} & \textbf{14.364} &
        \textbf{38.78} & \textbf{0.021} & \textbf{0.977} & \textbf{15.815} \\

        \bottomrule
    \end{tabular}
    }
    \label{tab:background_refinement_ablation}
\end{table}
\begin{table}[t]
    \centering
    \caption{Ablation study of disentangled-timestep sampling on foreground reconstruction.
    }
    \small
    \resizebox{0.8\linewidth}{!}{
    \begin{tabular}{l|cccc}
        \toprule
        Method &
        PSNR $\uparrow$ &
        SSIM $\uparrow$ &
        LPIPS$\downarrow$ &
        FID $\downarrow$ \\
        \midrule
        w/o Disentangled Sampling & 27.067 & 0.931 & 0.068 & 38.012 \\
        \rowcolor{blue!15}
        w/ Disentangled Sampling & \textbf{29.160} & \textbf{0.938} & \textbf{0.053} & \textbf{32.967} \\
        \bottomrule
    \end{tabular}
    }
    \label{tab:disentangled_sampling}
\end{table}
\begin{table}[h]
    \vspace{-12pt}
    \centering
    \caption{Dual-branch ablation on video layered object insertion.
    }
    \resizebox{0.95\linewidth}{!}{
        \begin{tabular}{l|c|ccccc|cc}
        \toprule
        Method (RGB-RGBA) &
        ViCLIP-T $\uparrow$ &
        \makecell{Background\\Consistency $\uparrow$} &
        \makecell{Subject\\Consistency $\uparrow$} &
        \makecell{Motion\\Smoothness $\uparrow$} &
        Aesthetic $\uparrow$ &
        \makecell{Imaging\\Quality $\uparrow$} &
        CLIP-I $\uparrow$ &
        DINO-I $\uparrow$ \\
        \midrule
        
        Single-branch (w/ DoRA)
        & 24.340 & 0.933 & 0.937 & 0.993 & 0.534 & 0.638 & 0.784 & 0.698 \\

        Single-branch (w/ LoRA)
        & 24.323 & 0.936 & 0.949 & 0.993 & 0.529 & 0.622 & 0.779 & 0.703 \\

        \cellcolor{blue!15}Dual-branch (ours)
        & \cellcolor{blue!15}\textbf{24.794} & \cellcolor{blue!15}\textbf{0.948} & \cellcolor{blue!15}\textbf{0.963} & \cellcolor{blue!15}\textbf{0.993} & \cellcolor{blue!15}\textbf{0.535} & \cellcolor{blue!15}\textbf{0.660} & \cellcolor{blue!15}\textbf{0.791} & \cellcolor{blue!15}\textbf{0.748} \\

        \bottomrule
        \end{tabular}
    }
    \label{tab:dual_branch}
\end{table}

\section{More Quantitative Ablation: Effect of the refined background dataset on Background layer Reconstruction}
\begin{figure*}[t]
    \centering
    \includegraphics[width=\textwidth]{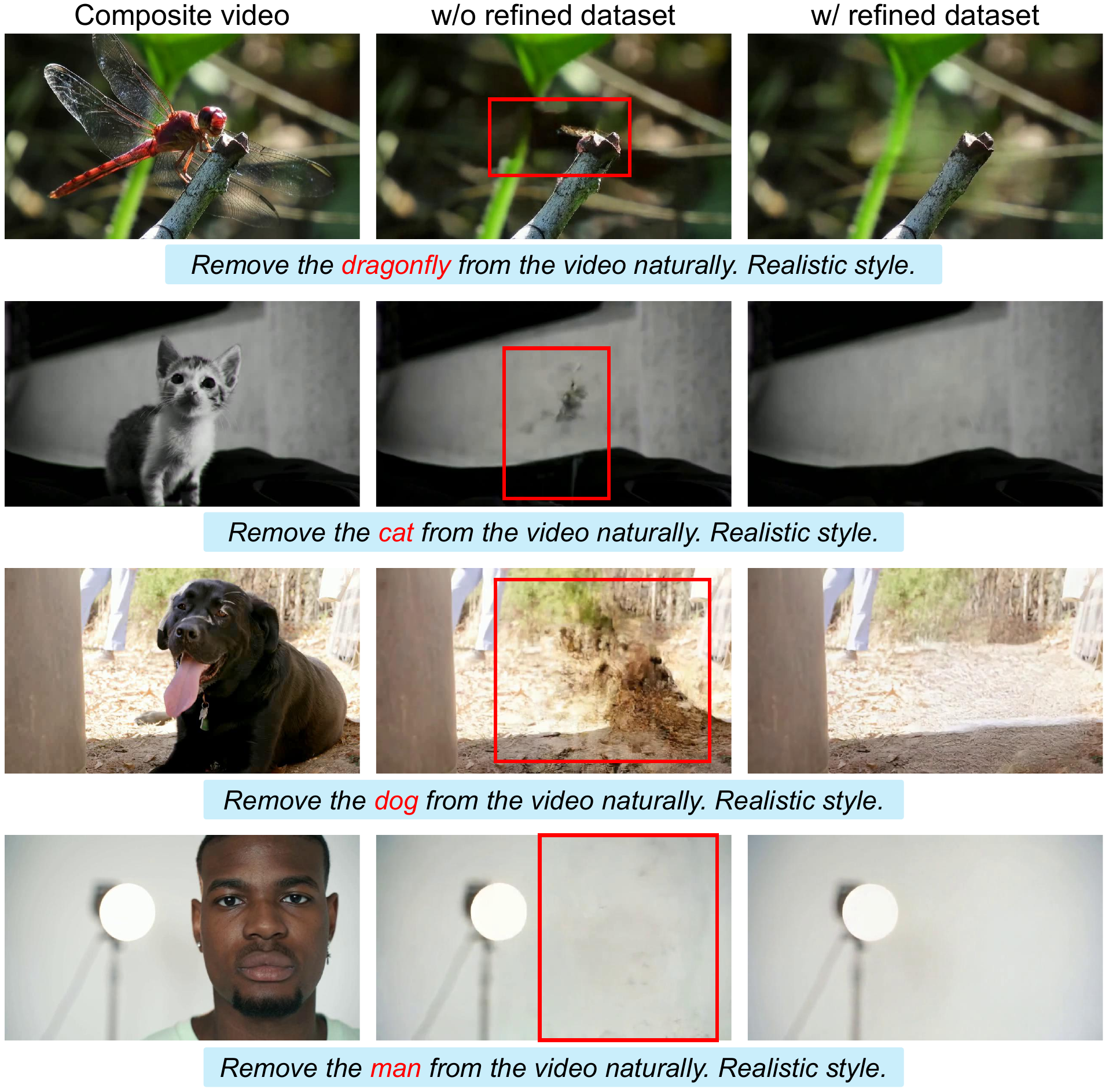}
    \vspace{-5ex}
    \caption{Qualitative ablation results demonstrating the effect of the refined background video dataset on background layer reconstruction.
    }
    \label{fig:more_agbi_effect}
\end{figure*}

We investigate the effect of using the refined background video dataset during training, where backgrounds from the object removal model are further refined using \MethodNameRefinement{}.
As shown in \cref{tab:background_refinement_ablation}, training with the refined background dataset yields slight improvements in background reconstruction performance across pixel-wise metrics and FID. Moreover, the visual improvements are more pronounced, as shown in \cref{fig:more_agbi_effect}, where models trained with the refined backgrounds produce substantially fewer artifacts and cleaner, more visually plausible reconstructions.
This demonstrates that artifacts from existing object-removal models can be propagated into the training process when their outputs are directly used as the background video dataset, causing the reconstruction model to inherit their limitations. 
By applying \MethodNameRefinement{} to refine the background videos, we provide cleaner supervision and consequently improve the quality of the reconstructed backgrounds. 

\section{More Quantitative Ablation: Disentangled Timestep Sampling}
Following UniCon~\citep{li2024simple}, we adopt disentangled-timestep sampling during inference to decouple the denoising trajectories of the RGB and RGBA branches while preserving cross-branch interaction.
This strategy alleviates optimization interference between the two branches, allowing each branch to focus on its own generation objective while still exchanging complementary information through joint attention.
As shown in \cref{tab:disentangled_sampling}, disentangled-timestep sampling consistently improves all evaluation metrics. These results indicate that disentangled-timestep sampling leads to more stable joint optimization and produces foreground layers with both higher reconstruction accuracy and better perceptual quality.

\begin{figure*}[t]
    \centering
    \includegraphics[width=0.95\textwidth]{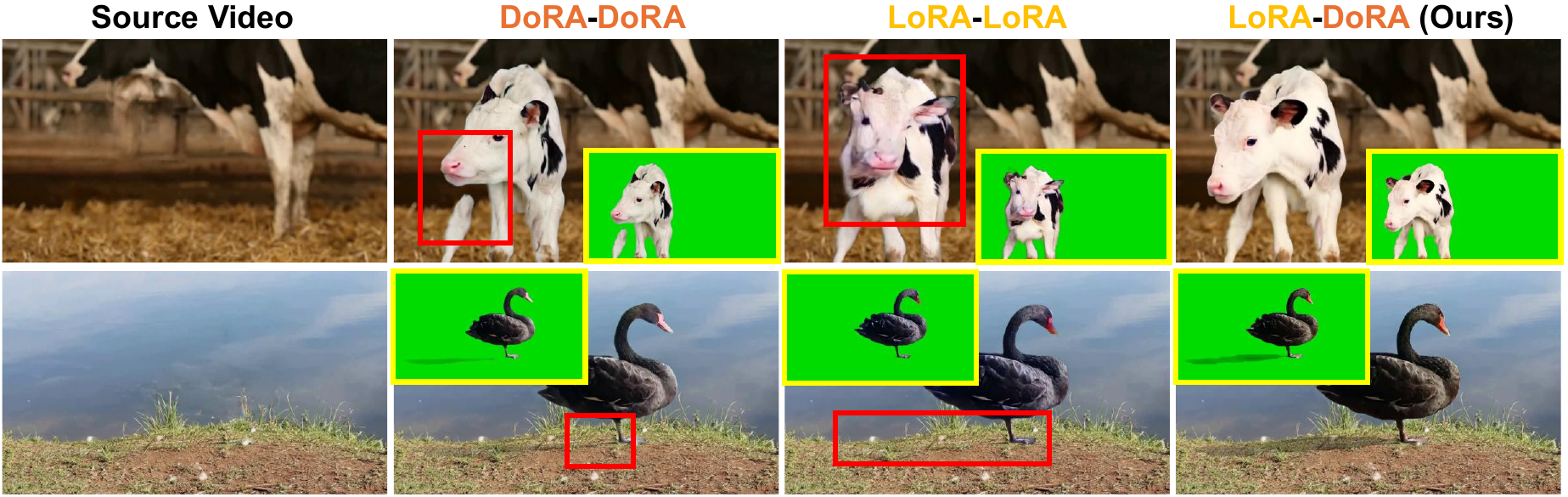}
    \vspace{-2.5ex}
    \caption{Qualitative ablation on the effect of lora-dora training strategy. The yellow box denotes the foreground layer.}
    \label{fig:lora_dora}
\end{figure*}

\begin{table}[h]
    \centering
    \caption{Foreground-conditioning ablation on foreground reconstruction.
    }
    \label{tab:fg_conditioning_ablation}
    \resizebox{0.7\linewidth}{!}{
    \begin{tabular}{lcccc}
        \toprule
        Method & PSNR~$\uparrow$ & SSIM~$\uparrow$ & LPIPS~$\downarrow$ & FID~$\downarrow$ \\
        \midrule
        w/o FG-conditioning & 24.876 & 0.919 & 0.072 & 40.105 \\
        \rowcolor{blue!15}
        w/ FG-conditioning & \textbf{29.160} & \textbf{0.938} & \textbf{0.053} & \textbf{32.967} \\
        \bottomrule
    \end{tabular}
    }
\end{table}
\begin{figure*}[t]
    \centering
    \includegraphics[width=0.75\textwidth]{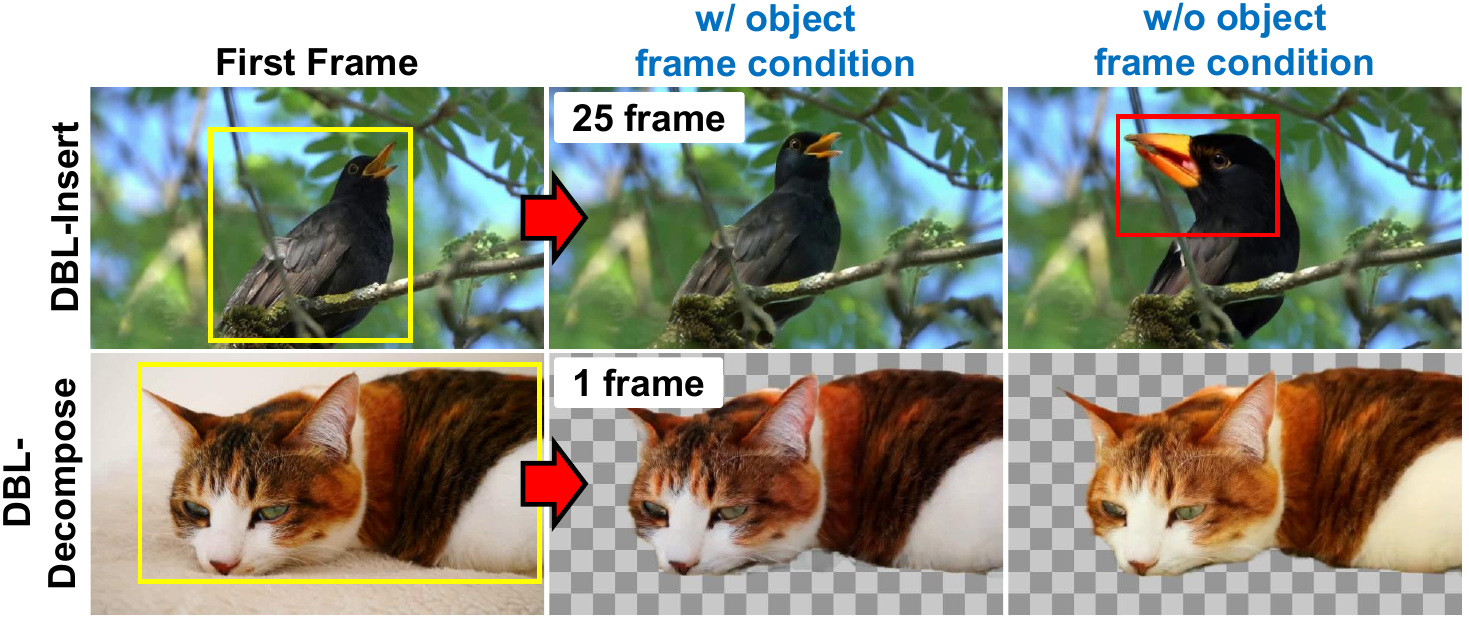}
    \vspace{-2ex}
    \caption{Qualitative ablation on the effect of foreground object conditioning. The yellow box denotes the input mask.}
    \label{fig:object_condition}
\end{figure*}

\section{More Qualitative Ablation: Dual-branch structure}

We compare single- and dual-branch structures to examine the effect of explicitly separating RGB and RGBA representations (\cref{tab:dual_branch} and \cref{fig:abl_single_branch}). In the single-branch structure, the RGB video and RGBA foreground latents are concatenated and jointly denoised in a shared representation space, after which each is decoded using its corresponding VAE.

The single-branch structure exhibits inconsistencies between the appearance of the inserted object in the edited first frame and the generated foreground layer. 
It also produces semi-transparent object regions, unstable visual effects, and noticeable visual artifacts. 
These failures can be attributed to the distribution gap between RGB and RGBA latents: jointly modeling them within a shared representation space makes it difficult for the single-branch model to sufficiently learn both representations. 
In contrast, our dual-branch structure explicitly separates the RGB and RGBA representations while allowing them to interact through joint cross-attention, producing high-quality and consistent foreground layers with more faithful object appearance and stable visual effects.

\section{More Qualitative Ablation: LoRA-DoRA strategy}
We compare LoRA-only, DoRA-only, and the hybrid LoRA--DoRA configuration
(\cref{fig:lora_dora}). Both LoRA-only and DoRA-only exhibit clear
failure cases, often producing unstable foreground layers or inconsistent
compositing. Because the two branches exchange information through
cross-attention, the single-method variants tend to show similar
degradations across both outputs. In practice, the hybrid LoRA--DoRA
configuration provides the most stable behavior, avoiding the severe
artifacts observed in the single-method variants. 

\section{More Qualitative Ablation: Foreground conditioning}
We analyze the role of the foreground conditioning videos $C_F$ and
$C'_F$ used in the RGBA branches of \MethodNameInsert{} and
\MethodNameDecompose{}. 
While both models can produce outputs without
these signals, the conditioning significantly improves visual quality
and temporal consistency, as shown in \cref{fig:object_condition} and
\cref{tab:fg_conditioning_ablation}. Quantitatively, foreground
conditioning substantially improves reconstruction quality, leading to
better pixel-wise metrics (PSNR/SSIM/LPIPS) and FID compared to the model
without foreground conditioning.
For \MethodNameInsert{}, the conditioning provides a high-quality appearance
anchor derived from the edited first frame, similar to the benefits
observed in hybrid image–video diffusion pipelines~\citep{kim2025videofrom3d}.
For \MethodNameDecompose{}, $C'_F$ offers a clearer cue for the object's
appearance, enabling more accurate RGBA layer recovery. Overall, the
foreground conditioning plays an important role in producing stable and
visually more coherent outputs.

\begin{figure*}[t]
    \centering
    \includegraphics[width=\textwidth]{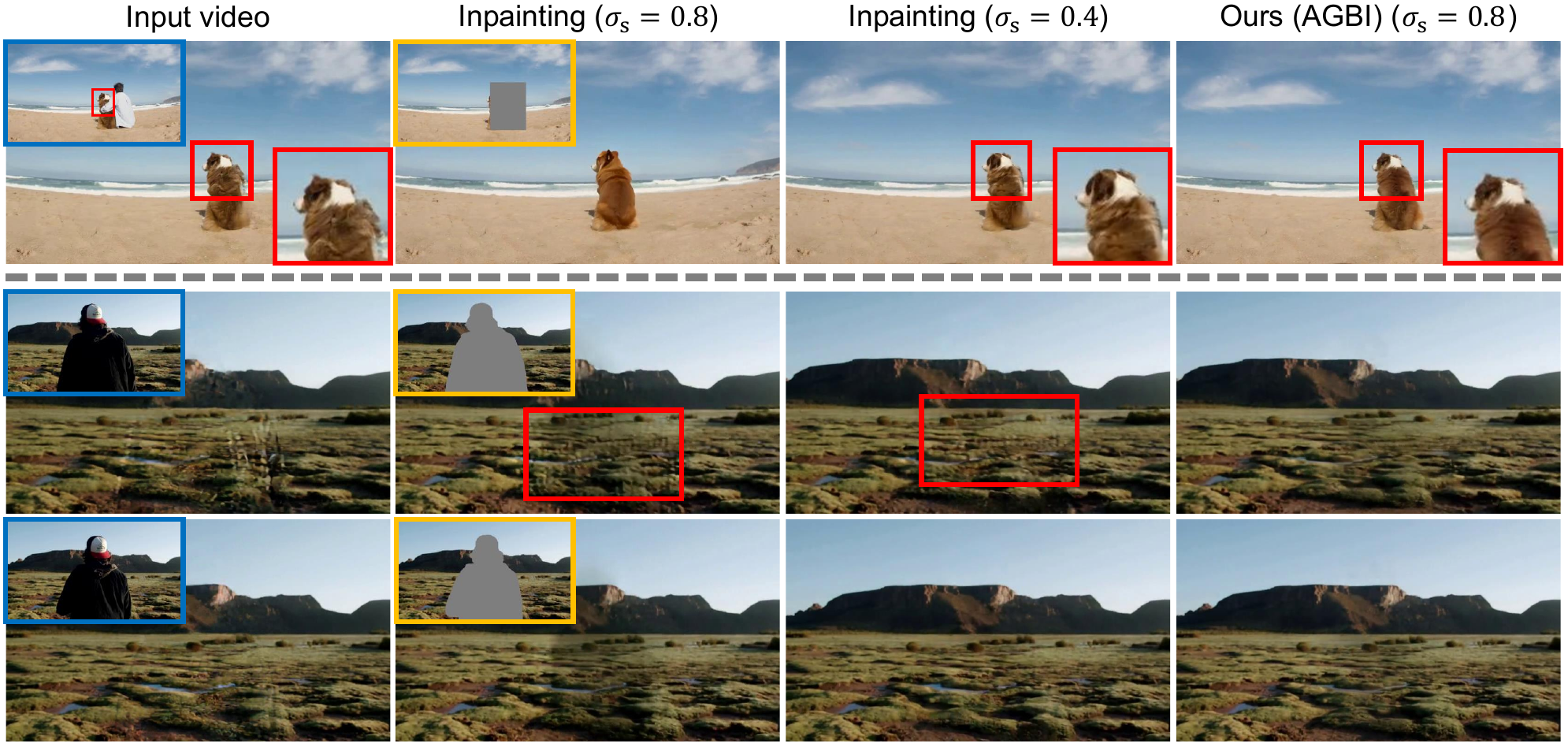}
    \vspace{-4ex}
    \caption{
    Comparison of background refinement with inpainting-based results.
    The input video is the source background video obtained from the object removal model and contains residual artifacts.
    The blue and orange boxes denote the original and masked video, respectively.
    }
    \label{fig:abl_AGBI_inpainting}
\end{figure*}
\begin{figure*}[t]
    \centering
    \includegraphics[width=\textwidth]{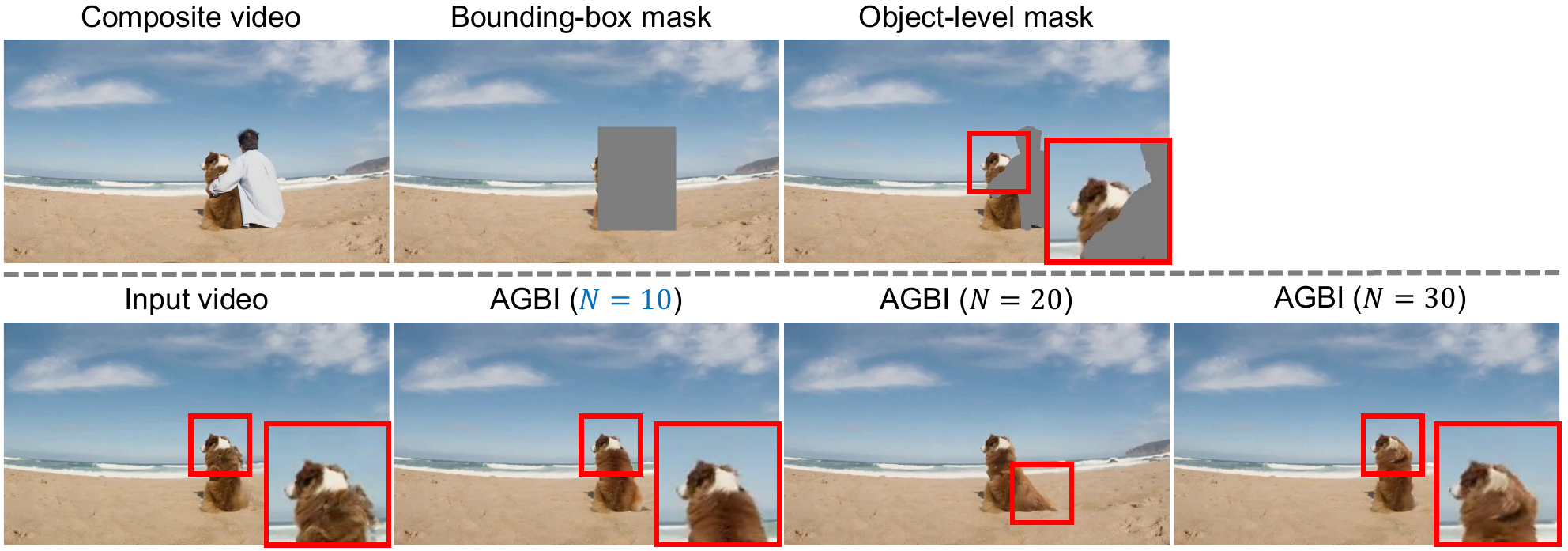}
    \vspace{-4ex}
    \caption{
    Effect of the number of initial injection steps ($N$).
    The input video is the source background video obtained from the object removal model and contains residual artifacts.
    }
    \label{fig:abl_AGBI_step}
\end{figure*}
\begin{figure*}[t]
    \centering
    \includegraphics[width=\textwidth]{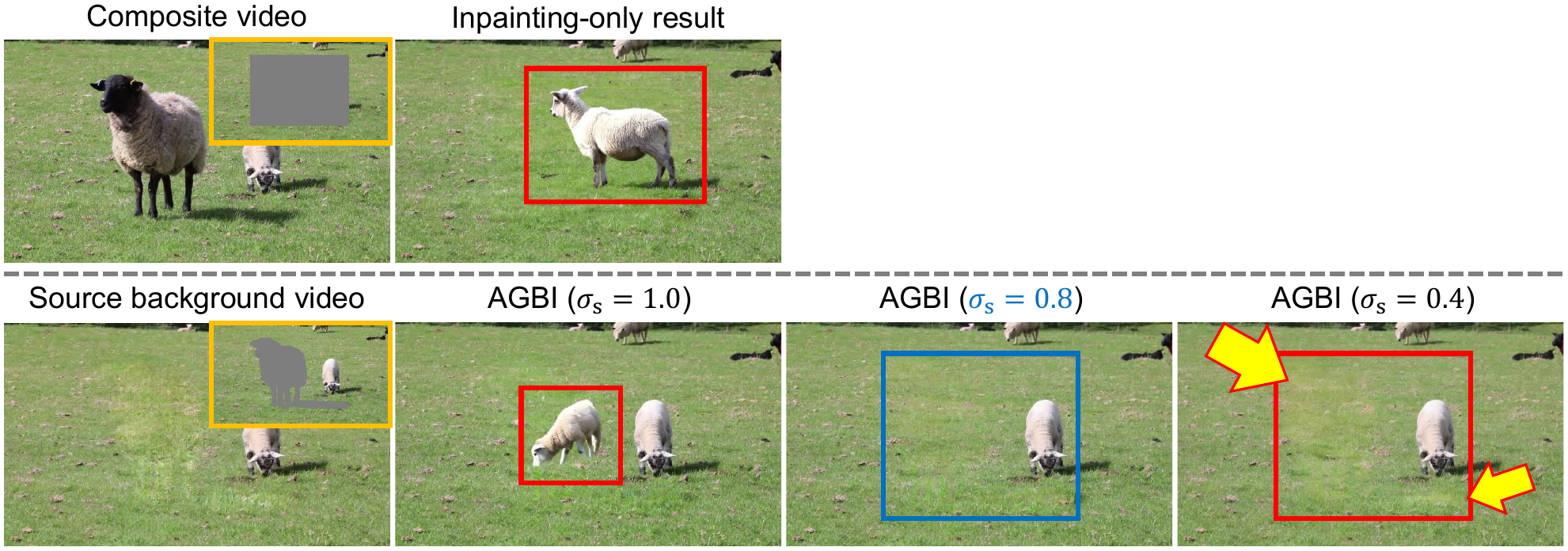}
    \vspace{-4ex}
    \caption{
    Effect of noise strength.
    The first row shows results obtained using inpainting with the bounding-box mask only.
    The source background video is obtained from the object removal model and contains residual artifacts.
    The orange box denotes the conditioning masked video, such as a bounding-box and object-level mask.
    }
    \label{fig:abl_AGBI_strength}
\end{figure*}
\begin{figure*}[t]
    \centering
    \includegraphics[width=\textwidth]{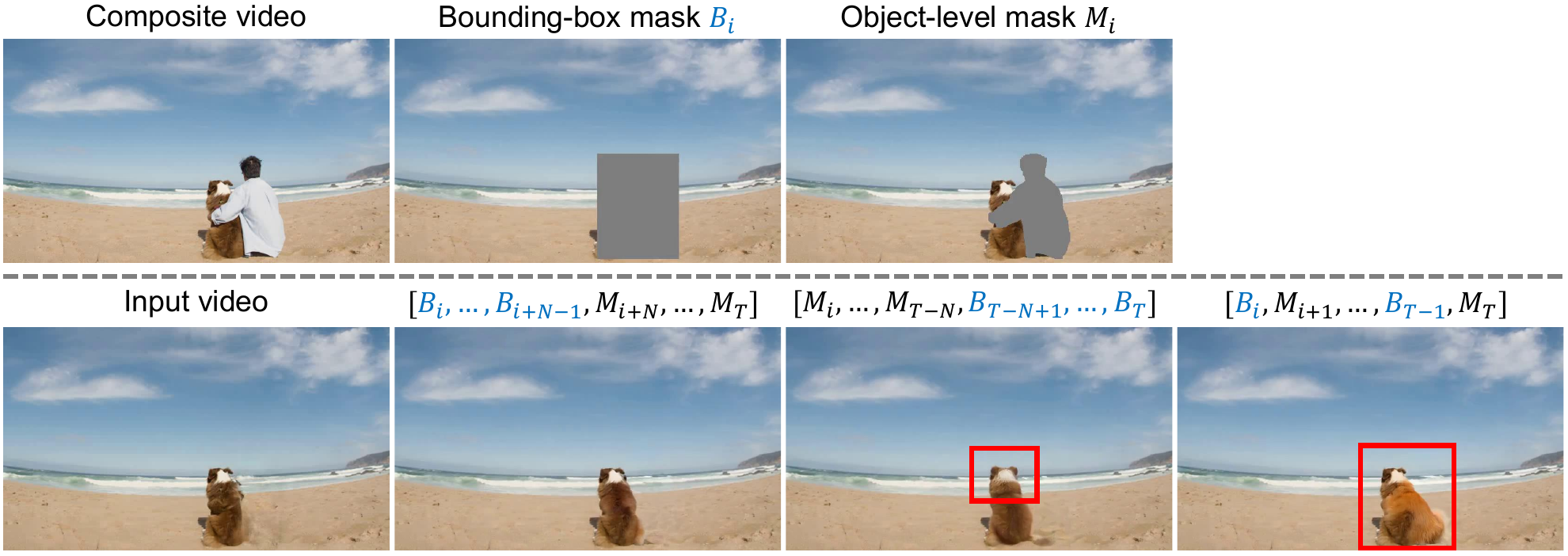}
    \vspace{-4ex}
    \caption{
    Effect of mask sequence order in \MethodNameRefinement{}.
    The input video is the source background video obtained from the object removal model and contains residual artifacts. 
    Three different mask sequence ordering variants are compared over the steps $i \in [T_s,T]$, 
    with $\sigma_s=0.8$ and $N=10$.
    }
    \label{fig:abl_AGBI_masking}
\end{figure*}

\section{More Qualitative Ablation: Background Refinement}
\begin{figure*}[t]
    \centering
    \includegraphics[width=\textwidth]{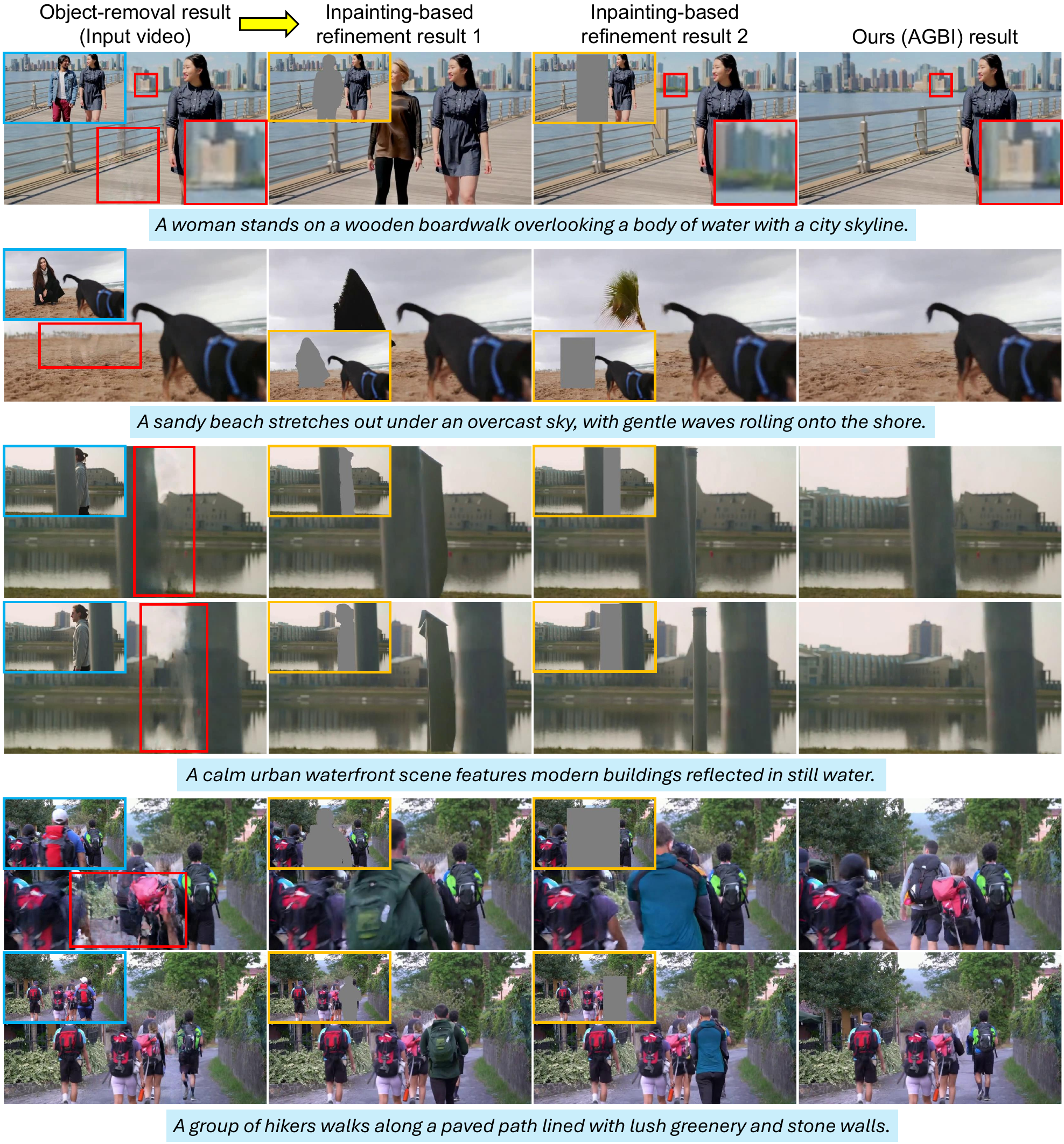}
    \vspace{-5ex}
    \caption{More comparison of background refinement with inpainting-based results.
    The input videos are background reconstruction results obtained using an object removal model. 
    The inpainting-based results use object-level and bounding-box masking, respectively. The last column shows our \MethodNameRefinement{} results. 
    The blue and orange boxes denote the original and masked video, respectively.
    }
    \label{fig:agbi_more_qual}
\end{figure*}

In all experiments, we follow the \MethodNameRefinement{} settings reported in Sec.~\ref{sec:AGBI}.
More qualitative comparisons of \MethodNameRefinement{} against inpainting-based background refinement methods are shown in \cref{fig:agbi_more_qual}.

\label{sec:AGBI_ablation}
\paragraph{Effect of two-stage masking strategy.}
To validate the effectiveness of our two-stage masking strategy, we compare
\MethodNameRefinement{} with two inpainting-based variants that use only
the object-level mask or the bounding-box mask, respectively.
For a fair comparison, we initialize all variants from the same source
background video $X^{\mathrm{src}}$ with the noise strength $\sigma_s$
as defined in \cref{eq:agbi_source_noise}, thereby
providing identical appearance guidance during inpainting.

As shown in \cref{fig:abl_AGBI_inpainting}, applying only the bounding-box mask effectively
mitigates residual artifacts, but tends to alter the appearance of the
original object even with a relatively low noise strength ($\sigma_s=0.4$).
In contrast, using only the object-level mask better preserves the
surrounding background appearance, but fails to sufficiently suppress
residual artifacts and can produce blurry results 
or unintended mask-shaped structures (e.g., human-like dark shadows) in some frames.

Our two-stage masking strategy combines the complementary benefits of the
two masks: the initial bounding-box refinement effectively removes
residual artifacts, while the subsequent object-level refinement
preserves the original background appearance and fine-grained details.
These results demonstrate the effectiveness of the proposed hybrid
masking strategy for achieving both artifact removal and appearance
preservation.

\paragraph{Effect of initial injection step.}
We analyze the effect of the number of initial injection steps
$N$ during which the bounding-box mask is applied. We fix the source
noise strength to $\sigma_s=0.8$ and evaluate three configurations,
$N\in\{10,20,30\}$, out of a total of 40 denoising steps (i.e., $[T_s, T]=[11,50]$).

As shown in \cref{fig:abl_AGBI_step}, using a larger number of initial bounding-box steps
($N=20$ or $N=30$) causes noticeable deviations from the original source
appearance and leads to unnatural structures.
This is because prolonged bounding-box refinement exposes a
broader region to the inpainting-prior, including areas that
do not require refinement, leading to unnecessary modifications of their
appearance and details.
In contrast, $N=10$ provides sufficient steps to
suppress residual artifacts while limiting unnecessary changes to the
surrounding region. We therefore adopt $N=10$ as the default setting.

\paragraph{Effect of noise strength.}
To investigate the effect of noise strength on appearance guidance,
we evaluate \MethodNameRefinement{} with different noise strengths
$\sigma_s \in \{1.0, 0.8, 0.4\}$. As the noise strength decreases, more appearance information from the
source background $X^{\mathrm{src}}$ is retained in the initial latent, allowing the
inpainting process to better preserve the original background appearance.

As shown in \cref{fig:abl_AGBI_strength}, when $\sigma_s=1.0$, the target latent is
initialized from pure noise $\epsilon$ and therefore contains no appearance
information from the source background. During the initial $N$ steps,
the bounding-box masking consequently relies primarily on the
inpainting-prior, resulting in outputs that combine the inpainting-only result. 
In contrast, a lower noise strength of $\sigma_s=0.4$ retains stronger
appearance information from the source background, but also leaves
residual artifacts such as afterimages and ghosting.

These results suggest that an appropriate noise strength is necessary
to balance artifact refinement and source appearance. 
Based on these experiments, we found $\sigma_s=0.8$ to provide a suitable balance between artifact removal and appearance preservation.

\paragraph{Effect of mask sequence order.}
We further investigate the effect of the mask sequence by considering
three variants: (1) applying the bounding-box mask $B$ during the
initial $N$ denoising steps, (2) applying the bounding-box mask $B$ during the later
$N$ steps, and (3) alternating between the bounding-box mask $B$ and the object-level
mask $M$ throughout denoising. Here, we follow the \MethodNameRefinement{} settings specified in \cref{tab:agbi_hyperparameters} (e.g., current denoising step $i \in[T_s, T]$).

As shown in \cref{fig:abl_AGBI_masking}, applying the bounding-box mask during the
later steps ($[M_i, \ldots, M_{T-N}, B_{T-N+1}, \ldots, B_{T}]$) leads to noticeable deviations from the original appearance, as the broader masked region unnecessarily alters regions that should remain unchanged. 
Therefore, the object-level mask is essential in the later steps for preserving the original appearance.
Alternating between the two masks ($[B_i, M_{i+1}, \ldots, B_{T-1}, M_{T}]$) also fails to consistently preserve the original appearance, since repeated application of the
bounding-box mask continues to overwrite appearance information during
denoising.

In contrast, our proposed ordering, which applies
$B$ during the initial steps followed by $M$ during
the later steps ($[B_i, \ldots, B_{i+N-1}, M_{i+N}, \ldots, M_{T}]$), 
first enables broad artifact suppression and then
restricts refinement to the object region. 
This allows the remaining artifacts to be refined in the later stages 
while preserving the appearance of the surrounding background.

\section{More qualitative comparisons: Video Layer Decomposition}
\begin{figure*}[t]
    \centering
    \includegraphics[width=\textwidth]{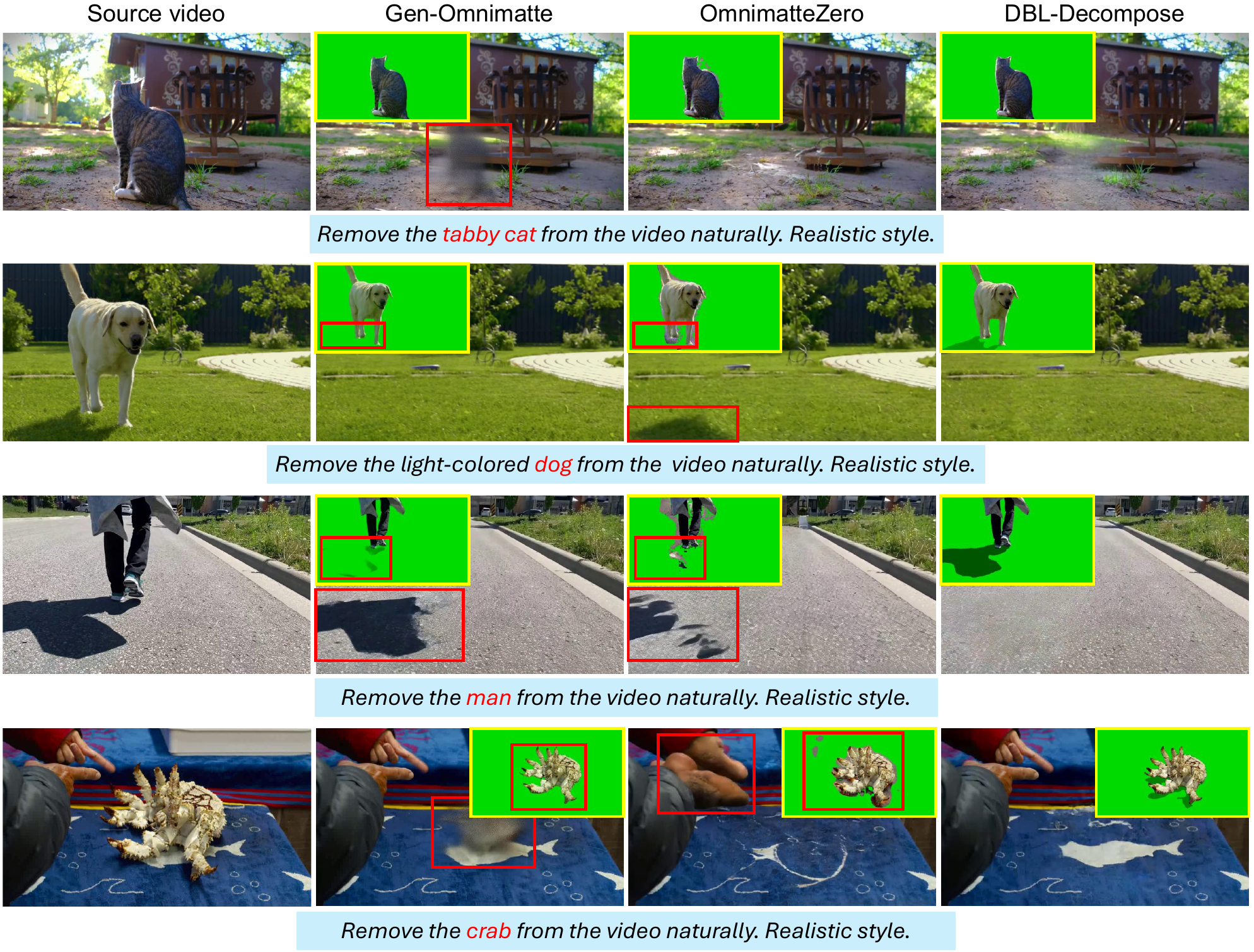}
    \vspace{-5ex}
    \caption{More qualitative video layer decomposition comparison with existing SOTA models. The yellow box denotes the foreground layer.
    }
    \label{fig:more_LD_comparison}
\end{figure*}

As discussed in Sec.~\ref{sec:benchmark_discussion}, no existing dataset provides ground-truth
foreground layers with object-induced visual effects (e.g., shadows, reflections) for
in-the-wild videos, which limits quantitative foreground reconstruction evaluation.
To complement the qualitative examples in \cref{fig:DBLDecomp_Qual} in the main paper,
as well as the FID metrics and human preference study reported in
\cref{tab:fid_comparison} and~\ref{tab:user_layer_decomposition}, 
we further analyze \MethodNameDecompose{} against existing video layer decomposition baselines using additional qualitative results on in-the-wild videos not used in constructing \DatasetName{}.
We provide additional qualitative comparisons against two representative decomposition baselines,
Gen-Omnimatte~\citep{lee2025generative} and OmnimatteZero~\citep{samuel2025omnimattezero},
in \cref{fig:more_LD_comparison}.

Both Gen-Omnimatte and OmnimatteZero often leave residual artifacts associated with the removed object in the background and incompletely capture object-induced visual effects in the foreground layer. Gen-Omnimatte additionally tends to produce semi-transparent foreground objects, while OmnimatteZero may produce less accurate object boundaries and noticeable changes in background appearance.
We attribute these failure cases to the absence of explicit, jointly supervised
foreground--background layer learning in these methods: Gen-Omnimatte relies on
per-video test-time optimization without paired ground-truth layers, while
OmnimatteZero is training-free and repurposes a pre-trained video diffusion model
without any layer-specific supervision. 

In contrast, \MethodNameDecompose{} is trained with explicit foreground layer supervision and
jointly models the RGB and RGBA branches through cross-branch attention, allowing
background and foreground reconstruction to remain mutually consistent. As shown in
\cref{fig:more_LD_comparison}, this enables DBL-Decompose to produce cleaner backgrounds
without residual artifacts and foreground layers that more completely and consistently
capture object appearance and associated visual effects.

\section{Failure Cases of \MethodName{}}
\begin{figure*}[t]
    \centering
    \includegraphics[width=\textwidth]{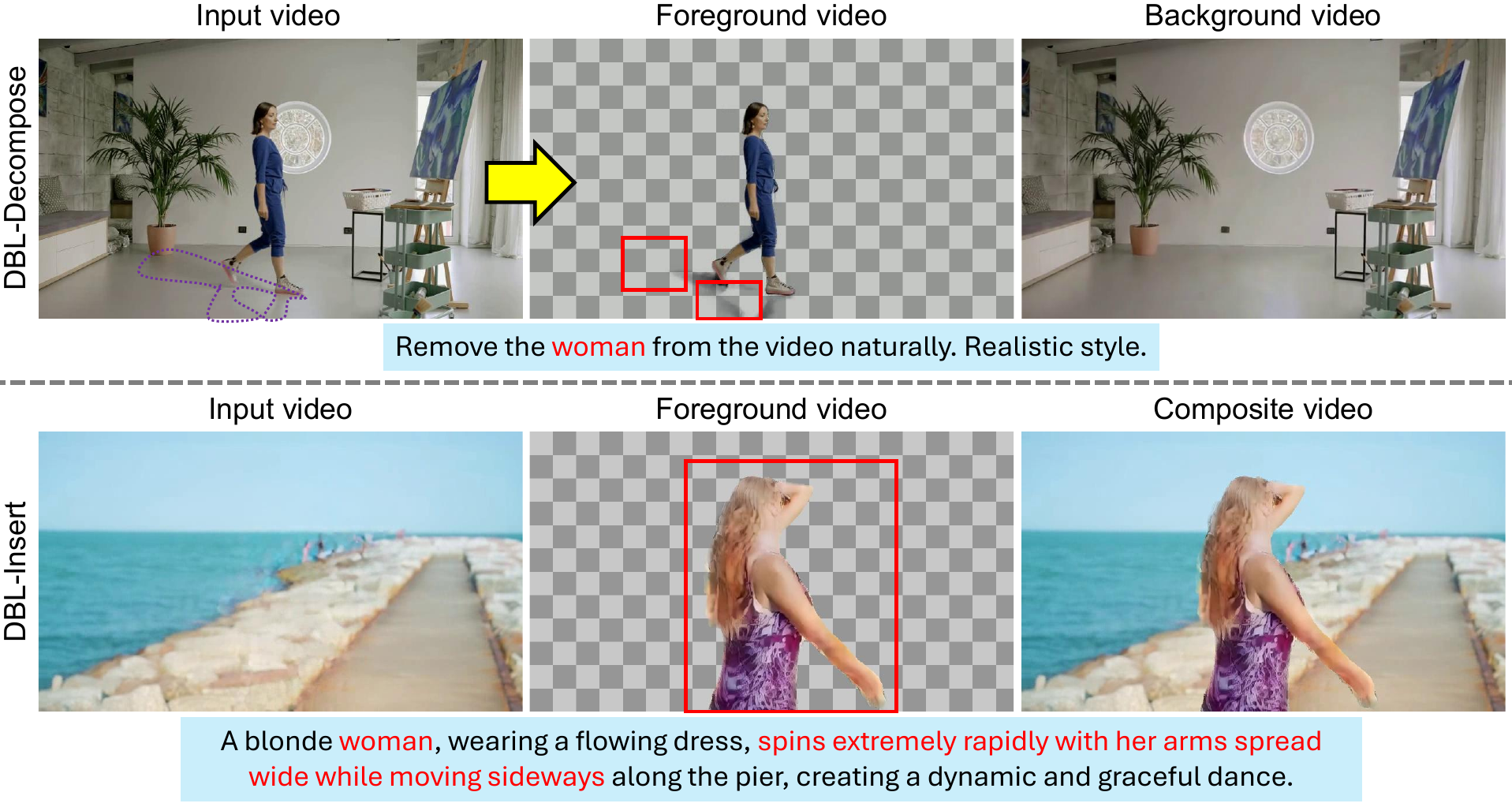}
    \vspace{-5ex}
    \caption{Failure cases of \MethodNameDecompose{} and \MethodNameInsert{}, respectively. The purple line denotes the complex visual effect.
    }
    \label{fig:failure_cases}
\end{figure*}

\cref{fig:failure_cases} illustrates representative failure cases of \MethodName{}. For \emph{\MethodNameDecompose{}}, the model may struggle with complex visual effects, particularly faint effects that are difficult to distinguish from the underlying surface, multiple overlapping effects, or object-induced effects that overlap with visual effects already present in the scene. Such cases can lead to incomplete separation of fine-grained or semi-transparent effects. Addressing these challenges remains an open research problem, highlighting the need for datasets that cover diverse and complex visual effects.

For \emph{\MethodNameInsert{}}, the model may struggle with complex or highly dynamic motions. Motions such as tumbling, spinning, or breakdancing can cause appearance artifacts during rapid object motion, leading to degraded fine-grained details. 
Addressing these failures remains an open research problem, closely related to recent efforts on controlling and generating videos with high-dynamic motions.

\section{More Qualitative Results}
In the main paper, we present only a single frame for each qualitative example due to space constraints. 
All qualitative examples are drawn from videos in LayeredVid-Benchmark and the manually created example. 
For better visualization of layer editing results, we crop selected regions to highlight key effects and perform composition using \emph{Adobe Premiere Pro}. All VACE-F and \MethodNameInsert{} results are obtained by compositing the foreground layer onto the background video.

\begin{figure*}[t]
    \centering
    \includegraphics[width=0.98\textwidth]{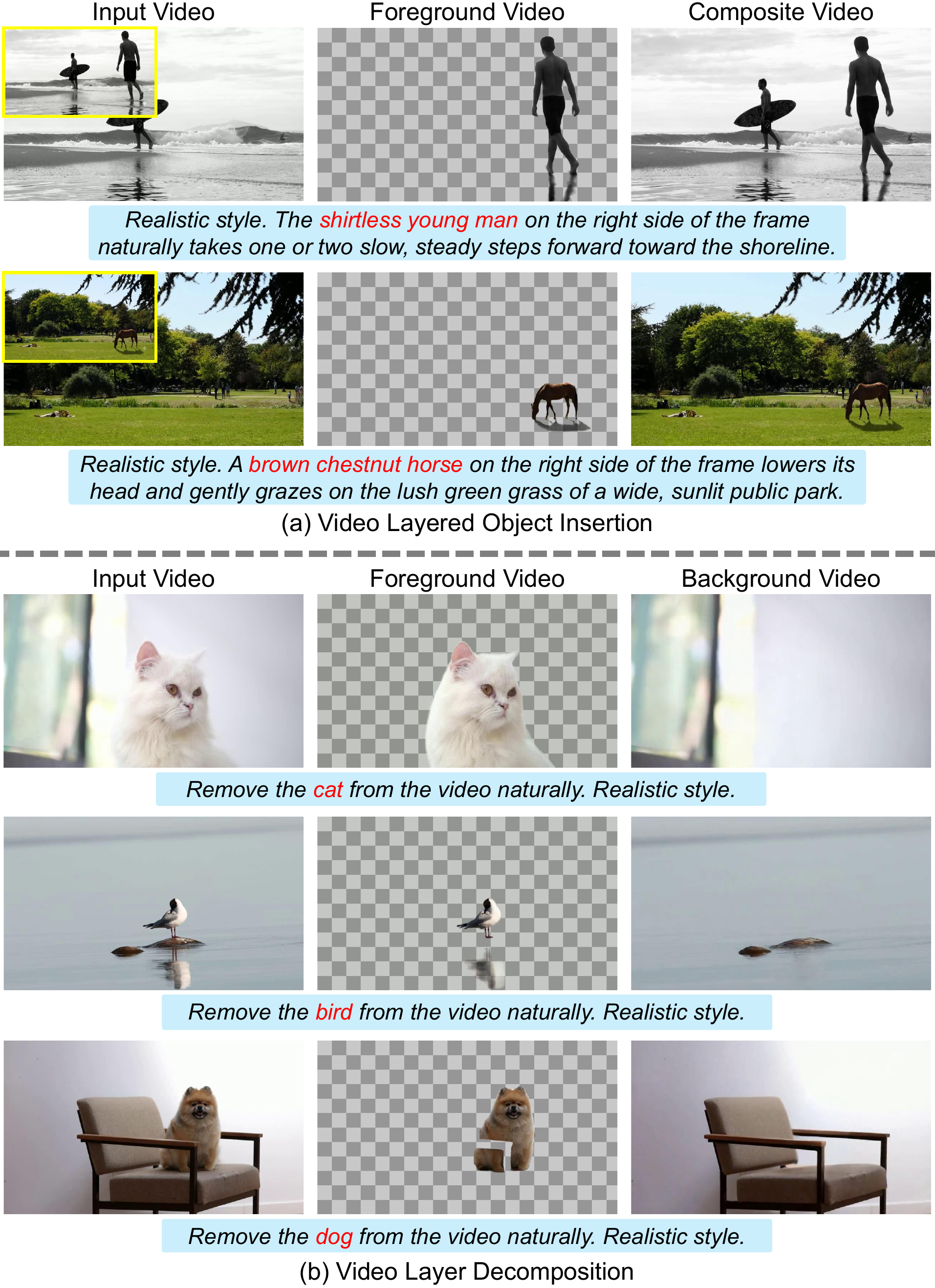}
    \vspace{-2.5ex}
    \caption{More qualitative examples: (a) Video layered object insertion using \MethodNameInsert{}. The yellow box denotes the edited first frame. (b) Video layer decomposition using \MethodNameDecompose{}.
    }
    \label{fig:more_qual_merge}
\end{figure*}

\begin{figure*}[t]
    \centering
    \includegraphics[width=\textwidth]{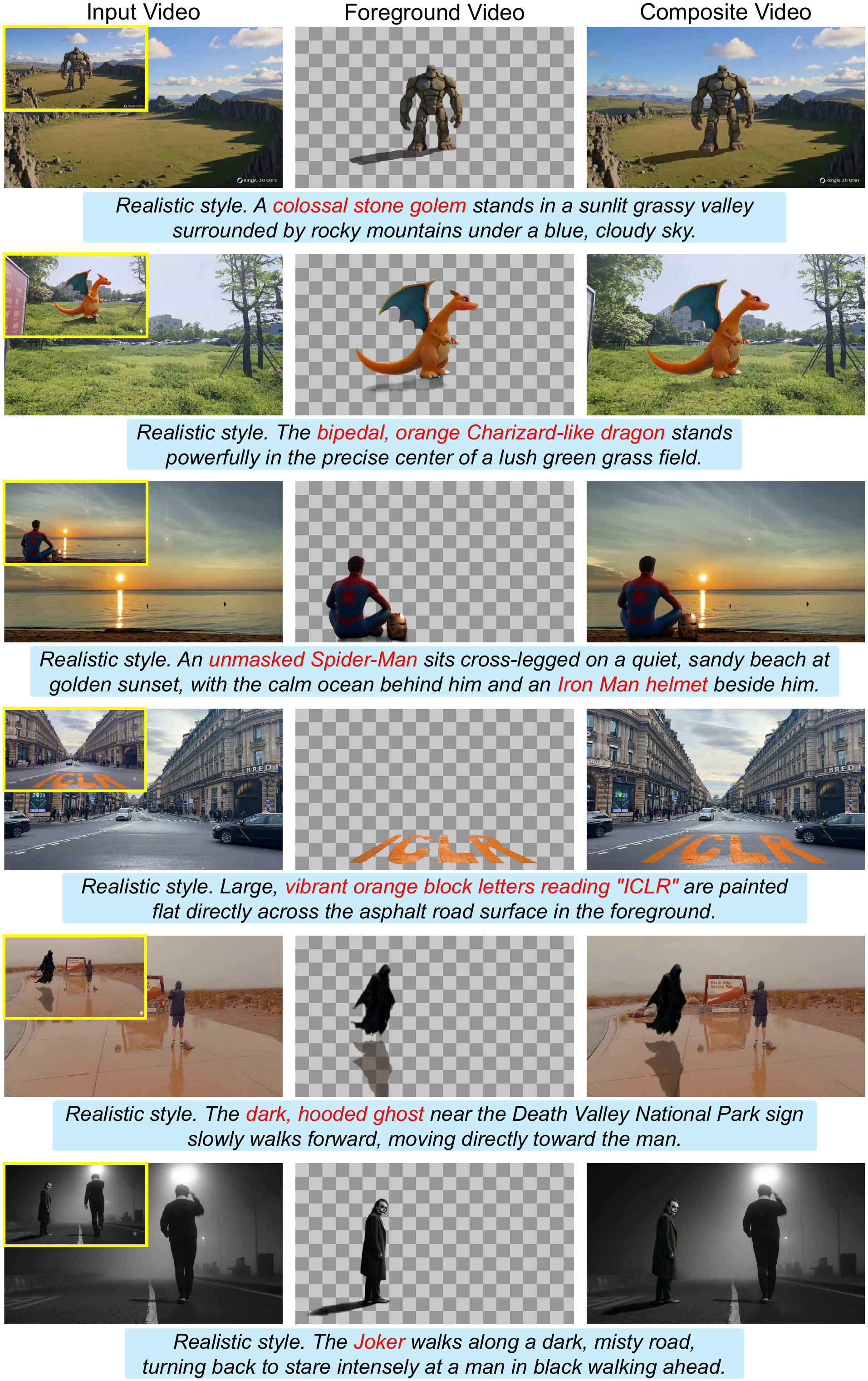}
    \vspace{-5ex}
    \caption{Qualitative results of \MethodNameInsert{} across diverse object types, demonstrating its robust video layered object insertion capability. The yellow box denotes the edited first frame.
    }
    \label{fig:more_various_insert}
\end{figure*}

More qualitative results for video layered object insertion and video layer decomposition are presented in \cref{fig:more_qual_merge}, respectively. 
Additionally, qualitative results demonstrating the robust performance of \MethodNameInsert{} across diverse object types are shown in \cref{fig:more_various_insert}. 
These examples demonstrate generalization beyond the realistic, object-centric scenarios represented in the \DatasetName{} dataset, including more diverse inputs such as unseen objects (e.g., golem, ghost, and text), animated characters, multiple objects, and movie-style characters.

Qualitative results are provided in the supplementary video.

\clearpage



\end{document}